\pdfoutput=1
\documentclass[11pt]{article}

\usepackage[letterpaper,margin=1in]{geometry}
\usepackage[T1]{fontenc}
\usepackage{newpxtext}
\usepackage{amsmath}
\usepackage{amssymb}
\usepackage{amsthm}   %
\usepackage{newpxmath}

\usepackage[dvipsnames]{xcolor}
\usepackage{graphicx}
\usepackage{booktabs}
\usepackage{xspace}
\usepackage[hyphens]{url}
\usepackage[numbers,sort&compress]{natbib}
\usepackage[format=plain,labelformat=simple,font=small]{caption}
\usepackage[shortlabels,inline]{enumitem}
\usepackage{wrapfig}
\usepackage{needspace}

\usepackage{enumitem}
\usepackage{booktabs}
\usepackage{multirow}
\usepackage{makecell}
\usepackage[flushleft]{threeparttable}
\usepackage{amsmath}
\usepackage{amssymb}
\usepackage{bm}
\usepackage{tabularx}
\usepackage{algorithm}
\usepackage{algpseudocode}

\usepackage{tikz}
\usetikzlibrary{arrows.meta,positioning,calc}

\graphicspath{{figures/}{./}}

\usepackage{colortbl}
\usepackage{pifont}
\definecolor{bestc}{RGB}{178,205,236}     %
\definecolor{secondc}{RGB}{219,230,247}    %
\newcommand{\cb}{\cellcolor{bestc}}        %
\newcommand{\cs}{\cellcolor{secondc}}      %
\newcommand{\yes}{\ding{51}}               %

\newcommand{\cA}{\mathcal{A}}
\newcommand{\cS}{\mathcal{S}}
\newcommand{\R}{\mathbb{R}}
\newcommand{\heldout}{\mathcal{H}}
\newcommand{\fte}{\mathrm{FTE}}
\newcommand{\spr}{\mathrm{SPR}}
\newcommand{\wh}{\widehat}
\newcommand{\bx}{\mathbf{x}}
\newcommand{\bz}{\mathbf{z}}
\newcommand{\bphi}{\boldsymbol{\phi}}

\DeclareMathOperator*{\argmax}{arg\,max}

\usepackage{placeins}
\usepackage{amsthm}
\newtheorem{theorem}{Theorem}
\newtheorem{proposition}{Proposition}
\newtheorem{lemma}{Lemma}

\DeclareMathOperator*{\Cov}{Cov}

\usepackage[pagebackref,breaklinks,colorlinks,citecolor=blue,linkcolor=red,urlcolor=magenta]{hyperref}
\makeatletter
\newcommand{\blfootnote}[1]{%
  \begingroup
  \renewcommand{\thefootnote}{}%
  \renewcommand{\@makefntext}[1]{\parindent\z@\noindent##1}%
  \footnote{\raggedright #1}%
  \addtocounter{footnote}{-1}%
  \endgroup
}
\makeatother

\newcommand{\papertitle}{RiPPLE: Cross-Space Performance Prediction from Early
Training for Neural Architecture Search}
\title{\papertitle}

\author{%
Yifan Yang\textsuperscript{1*} \quad
Zhaoyan Wang\textsuperscript{2*} \quad
Zheng Gao\textsuperscript{1}\\[0.35em]
Xiaoyu Li\textsuperscript{1} \quad
Jiaojiao Jiang\textsuperscript{1}\\[0.7em]
{\small\textsuperscript{1}University of New South Wales}\\[0.2em]
{\small\textsuperscript{2}Korea Advanced Institute of Science and Technology}}

\date{}
\hypersetup{%
  pdftitle={RiPPLE: Cross-Space Performance Prediction from Early Training for Neural Architecture Search},
  pdfauthor={Yifan Yang, Zhaoyan Wang, Zheng Gao, Xiaoyu Li, Jiaojiao Jiang},
  pdfsubject={Neural architecture search and performance prediction},
  pdfkeywords={neural architecture search, performance prediction, machine learning}
}

\newcommand{\maketitlesupplementary}{%
  \begin{center}
    {\Large\bfseries \papertitle}\\[0.6em]
    {\Large Supplementary Material}
  \end{center}
  \vspace{1.0em}
  \suppressfloats[t]%
}

\begin{document}
\maketitle
\blfootnote{\textsuperscript{*}These authors contributed equally to this work.\\
Corresponding author: Yifan Yang
({\hypersetup{urlcolor=black}\href{mailto:evanyifanyang2026@gmail.com}{\texttt{\bfseries evanyifanyang2026@gmail.com}}}).}
\begin{abstract}
Neural architecture search (NAS) evaluates candidate networks, and the most reliable way to measure one is
to train it fully, which is expensive, so evaluating an entire space this way is prohibitive. Two
cheaper families each fall short. Zero-cost proxies (ZCP) score every architecture at initialization, but
no single proxy holds across spaces, and even the strongest leaves a wide gap from a faithful
ranking. Fully supervised predictors are more accurate, yet every label is a full training of a sampled
architecture, and those adding a partial-training signal read it as a per-candidate
feature, so cost grows with the space. Our goal is to rank an entire space well at a fractional
budget. We introduce \textbf{RiPPLE}, \underline{R}anking v\underline{i}a
\underline{P}refix-\underline{P}ropagated \underline{L}abel
\underline{E}xtrapolation, which obtains its labels cheaply: it trains a small coverage set
of anchors to an early prefix, extrapolates each learning curve to a surrogate
label, and propagates these labels over label-free features. The signal stays a label on the
anchors, never a per-candidate feature, and no per-space choice reads held-out accuracy, so one
unchanged pipeline ranks the spaces at fractional budgets. Across twelve cells from four
search-space families, RiPPLE ranks near the oracle, matching or surpassing predictors under a common budget ceiling
and holding where ZCP collapse; read as a selector, its chosen architectures are
competitive with leading search.
\end{abstract}

\section{Introduction}
\label{sec:intro}

\begin{figure}[t]
  \centering
  \setlength{\abovecaptionskip}{3pt}
  \includegraphics[width=0.62\textwidth]{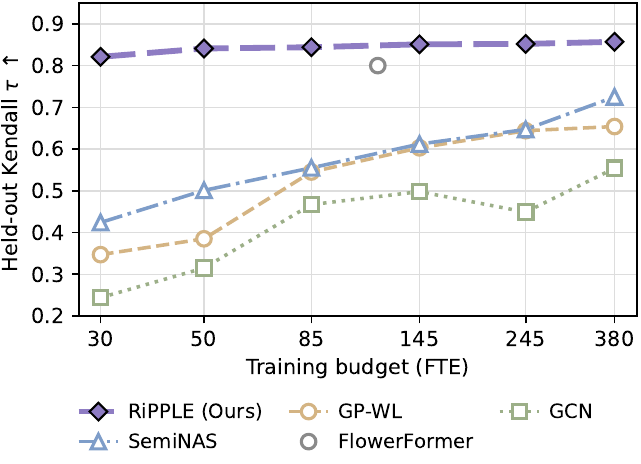}
  \caption{Ranking quality versus reported training budget on NAS-Bench-201/CIFAR-10.
  RiPPLE is shown with GP-WL~\cite{ru2021nasbowl}, GCN~\cite{dudziak2020brpnas},
  SemiNAS~\cite{luo2020seminas}, and the external FlowerFormer~\cite{hwang2024flowerformer}
  reference. FlowerFormer's reported label budget and independent-test protocol are
  not harmonized with RiPPLE's FTE accounting. The curves are descriptive comparisons,
  not a demonstration of strictly matched evaluation sets and total costs.}
  \label{fig:teaser}
\end{figure}

Neural architecture search (NAS) evaluates candidate networks by training them, and training is
expensive, so most of a search's cost is spent on architectures that are later
discarded~\cite{white2023nassurvey}. Performance predictors reduce this cost by learning to
rank architectures from a few measurements, and the ideal predictor would order the entire
space while training only a small part of it~\cite{white2021powerful}.

Two common predictor designs sit at opposite ends of this trade. Zero-cost proxies (ZCP) read a single
untrained pass and rank the whole space for free, but their agreement with true performance is
capped near a Spearman correlation of $0.9$ and degrades sharply across
spaces~\cite{abdelfattah2021zerocost}. Fully supervised predictors reach higher
agreement~\cite{dudziak2020brpnas,ning2020gates,yi2023narformer}, but each of their training
labels is a full training, while methods such as OMNI keep the partial-training signal
as a feature~\cite{white2021powerful}, so scoring a new architecture requires training it and the
cost grows with the number of candidates scored. Curve-extrapolation and multi-fidelity methods
share this property, and they return an early-stopping decision or a single configuration rather
than a ranking. Ranking the whole space accurately at a fractional training budget remains
challenging: the proxies cheap enough to reach it are weak, and the supervised predictors pay a
full training for every label.

Our key observation is that the partial-training signal need not be a feature. A feature must
be available for every architecture the predictor scores, whereas a label is needed only for
the few architectures it trains. If the signal enters as a label, the training cost is confined
to a small set of anchors, and the rest of the space is ranked from features that cost nothing
to compute. This decouples the training budget, which scales with the anchors, from the ranking
reach, which spans the whole space. The choice of where the signal enters is not a matter of
taste; it is the difference between a predictor that ranks a few thousand candidates and one
that ranks a few hundred million.

We build this observation into RiPPLE, a predictor that ranks an entire search space at a
fractional training budget. RiPPLE selects a few hundred anchors once by coverage, trains each
to a short prefix of its schedule, extrapolates the resulting curve into a surrogate label, and
propagates these labels across the space with a regressor over label-free features. The features
are a bank of zero-cost proxies together with an architecture encoding, both computable without
training any candidate, so the only training cost is the short prefix spent on the anchors.
None of the three components that could be tuned per benchmark reads a label: the proxy set and
the encoding each take all the candidates available for a space and let the propagator
down-weight the uninformative ones, and the curve reader follows a rule keyed to the schedule
shape that is fixed before any training. RiPPLE therefore carries no per-benchmark
hyperparameter search and transfers to a new space unchanged.

Under a ceiling of about $150$ full-training equivalents, one unchanged RiPPLE pipeline ranks
the whole space near the oracle in every one of twelve cells spanning four search-space
families, where no zero-cost proxy holds and the graph and sequence predictors that do well on
image cells are undefined on the size, recurrent, and transfer spaces. On NAS-Bench-201 it reaches a Kendall
$\tau$ of $0.85$ at the first tested point of $127$ equivalents (Table~\ref{tab:frontier});
published predictor results provide protocol-specific reference points rather than a
controlled label-cost ratio. Read as a selector, its top-ranked network lies within $0.08$ on CIFAR-10 and $0.26$ on CIFAR-100 of
the optima and is competitive with trained search on the DARTS space of
roughly $10^{18}$ architectures.

\noindent Our contributions are as follows.
\begin{itemize}[leftmargin=1.4em,itemsep=2pt,topsep=2pt]
  \item We propose RiPPLE, a predictor that spends its partial-training signal as a label on a few
  coverage anchors rather than a per-candidate feature, ranking an entire search space at a
  fractional training budget.
  \item We show that one unchanged pipeline ranks near the oracle across twelve cells from four
  search-space families, where no zero-cost proxy holds and no graph- or sequence-based predictor applies.
  \item We demonstrate that RiPPLE matches supervised predictors at a fraction of their labels while
  selecting near-optimal architectures.
\end{itemize}

\begin{figure}[t]
\centering
\includegraphics[width=\textwidth]{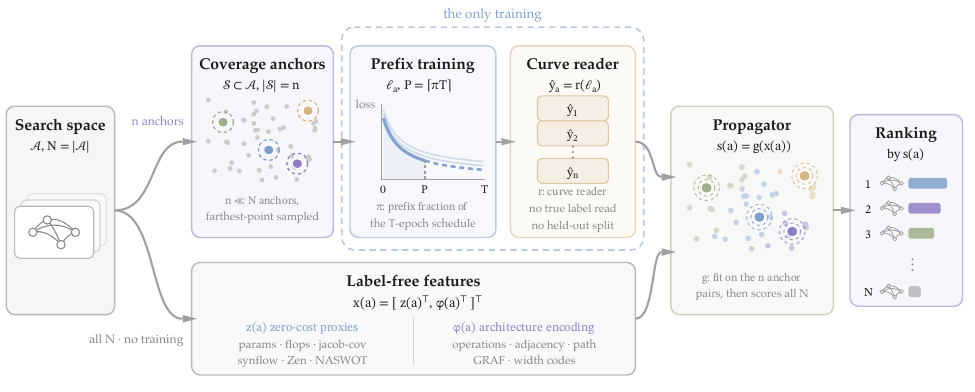}
\caption{RiPPLE ranks the whole space $\cA$ at a fractional budget. Only the $n$ coverage anchors
$\cS$ are trained, each to a short prefix that yields a learning curve $\ell_a$, which the reader
$r$ turns into an extrapolated label $\wh{y}_a$. A propagator $g$ is fit to predict these anchor
labels from label-free features $\bx=[\bz;\bphi]$, a bank of zero-cost proxies $\bz$ and an
architecture encoding $\bphi$ computed for every architecture; it then scores the whole space,
$s(a)=g(\bx(a))$, and $\cA$ is ranked by $s$. The partial-training signal thus enters as the anchor
label, not as a per-candidate feature, so the training cost scales with the anchors while the
ranking spans all of $\cA$.}
\label{fig:framework}
\end{figure}
\section{Related Work}
\label{sec:related}

Neural architecture search selects an architecture from a predefined space, and its search
strategies fall into three families: one-shot methods that share weights across a
supernet~\cite{pham2018enas,guo2020spos}, differentiable methods that relax the discrete
choice~\cite{liu2019darts}, and predictor-based methods that learn to rank architectures from a
sample of evaluations~\cite{white2023nassurvey}. The first two return a single architecture, whereas
a predictor ranks the whole space. RiPPLE is a predictor, so we review the low-cost signals a ranker
can consume and where each of them places the signal.

\noindent\textbf{Zero-cost proxies.} Zero-cost proxies score an architecture from a single forward
pass at initialization~\cite{abdelfattah2021zerocost}, and later work strengthens or assembles them,
as in ZiCo~\cite{li2021zico}, AZ-NAS~\cite{lee2024aznas}, and L-SWAG~\cite{casarin2025lswag}. Their
agreement with true accuracy stays capped near a Spearman correlation of $0.9$ on NAS-Bench-201 and
does not transfer across spaces~\cite{krishnakumar2022benchsuitezero,peng2024swap}. Because a proxy
is read before any training, it is a reference frontier a paid ranker must clear rather than an
equal-budget competitor, and a method that pretrains a predictor to reproduce a proxy ranking
inherits that quality for free. RiPPLE uses a bank of such proxies only as input features and reports
the proxy-only frontier separately.

\noindent\textbf{Learned performance predictors.} Supervised predictors fit an
architecture-to-accuracy map from fully trained examples~\cite{white2021powerful}.
BRP-NAS~\cite{dudziak2020brpnas} predicts with a graph convolutional network,
NAS-BOWL~\cite{ru2021nasbowl} fits a Gaussian process with a Weisfeiler-Leman graph kernel, and
SemiNAS~\cite{luo2020seminas} trains an LSTM encoder with self-training on unlabeled architectures.
Under a common ceiling of $150$ full-training equivalents, RiPPLE-prefix ranks above these baselines
on eleven of twelve cells. Unlike these fully supervised predictors, RiPPLE-prefix does not train an architecture to completion: its
labels are surrogate accuracies extrapolated from early-stopped prefixes, so each label costs a
fraction of a full training. A second line encodes the architecture for a transformer:
TNASP~\cite{lu2021tnasp} adds a Laplacian to the operation embeddings,
NAR-Former~\cite{yi2023narformer} tokenizes operations and connections,
NAR-Former~V2~\cite{yi2023narformerv2} adds a graph-aided block, NN-Former~\cite{xu2025nnformer} adds
sibling attention, and FlowerFormer~\cite{hwang2024flowerformer} passes messages along the data flow.
These encoders are accurate but label-hungry: the strongest reach a Kendall $\tau$ of $0.85$ on
NAS-Bench-201 only at several hundred labels, several never reach it, and all are scored on the whole
benchmark including their training pool, except for studies such as FlowerFormer that use an independent test set. Their advertised training budgets also omit a validation set of tens
to a few hundred fully trained architectures used for early stopping and tuning, which at the
smallest settings exceeds the counted training budget itself, whereas RiPPLE reads no held-out
accuracy and needs no such set. They are also developed on cell-structured image benchmarks and do
not define an encoding for the size or recurrent spaces. Gated predictors such as
GATES~\cite{ning2020gates} and its training-analogous extension TA-GATES~\cite{ning2022tagates} report
Kendall $\tau$ natively but share this label cost. Among feature-augmented predictors, OMNI~\cite{white2021powerful} augments an
encoding with a learning-curve statistic, but keeps that statistic as a per-candidate feature, so
scoring a candidate requires partially training it and the cost scales with the candidate count,
which its authors report as a barrier on large spaces. RiPPLE reads the same curve but places it on a
few hundred coverage anchors as a label, so its training cost is fixed while the ranking spans
hundreds of millions of architectures.

\noindent\textbf{Curve extrapolation and multi-fidelity search.} A separate line reads a partial
learning curve to stop early or steer a search. Domhan et al.~\cite{domhan2015speeding} extrapolate a
parametric curve; Hyperband and BOHB allocate a budget across configurations by successive
halving~\cite{li2018hyperband,falkner2018bohb}; LC-PFN scores configurations inside a freeze-thaw
loop~\cite{adriaensen2023ifbo}; and MPENAS~\cite{xu2023mpenas} fuses an encoding, a proxy, and a
curve into a cross-fidelity fitness. All of these partially train every candidate they score and
return a surviving configuration rather than a ranking. Weight-sharing supernets produce a
whole-space signal at one training cost but rank architectures
weakly~\cite{pham2018enas,liu2019darts,guo2020spos,yang2024pada} and bind to a single predefined
space. RiPPLE instead spends its budget on prefix-training a fixed set of anchors and propagates once
to the whole space, returning a whole-space ranking rather than a single surviving configuration,
unchanged across topological, size, and sequence spaces.

CoRA-NAS~\cite{yang2026cora} also uses early learning curves on a small anchor set to
refine a whole-space ranking without fully trained accuracy labels. It combines a coarse
zero-cost ranking prior with a learned residual correction from prior-stratified anchors.
RiPPLE instead directly propagates prefix-derived surrogate labels from encoding-space coverage anchors, using zero-cost
proxies as input features rather than as a separately corrected ranking prior.

\section{Method}
\label{sec:method}

RiPPLE ranks an entire search space at a fractional training budget in four stages, shown in
Fig.~\ref{fig:framework}. It computes label-free descriptors for every architecture, selects a small
anchor set by coverage, trains only the anchors to a short prefix and reads their curves into
extrapolated labels, and fits a propagator that carries those labels to the whole space.
Section~\ref{sec:method-setup} fixes the notation and the cost model, and
Section~\ref{sec:method-pipeline} details the four stages.

\subsection{Problem setup and notation}
\label{sec:method-setup}

We study predictor-based NAS. A search space $\cA$ contains $N=|\cA|$ candidate architectures, and
$y\colon\cA\to\R$ is the true performance of an architecture after full
training~\cite{white2021powerful}, oriented so that higher is better (e.g.\ accuracy), so standard
search seeks $a^{\star}=\argmax_{a\in\cA} y(a)$. Evaluating $y$ trains a network from scratch, so we instead build
a score $s\colon\cA\to\R$ whose induced ranking agrees with that of $y$ while training only a small
fraction of the space. Only the ranking induced by $s$ is used, so $s$ need not be calibrated
to $y$. We measure ranking quality by rank correlation on a held-out set $\heldout$ of architectures
the predictor never trains on.

\noindent\textbf{Cost model.} Training dominates cost, so we measure budget in full-training equivalents.
One $\fte$ is the cost of training a single architecture to the end of its schedule of $T$ epochs, and
training $n$ architectures each to a prefix $\pi\in(0,1]$ of the schedule spends
\begin{equation}\label{eq:fte}
  \fte = n\,\pi .
\end{equation}
This places fractional-training and full-training predictors on a single budget axis.

\noindent\textbf{Label-free features.} Each architecture $a$ carries two label-free feature vectors: a
structural encoding $\bphi(a)\in\R^{d}$ and a vector of zero-cost proxies $\bz(a)\in\R^{m}$. RiPPLE
takes the full standard proxy
bank~\cite{krishnakumar2022benchsuitezero,abdelfattah2021zerocost,li2021zico,jiang2023meco,peng2024swap}
and every encoding available for the architecture type, with no per-space selection, and lets the
propagator down-weight the uninformative coordinates. Concatenating the two gives
\begin{equation}\label{eq:feature}
  \bx(a) = \bigl[\,\bz(a)^\top,\ \bphi(a)^\top\,\bigr]^\top \in \R^{m+d},
\end{equation}
computed for the whole space without training any candidate, so that only the anchors incur training.

\subsection{The RiPPLE framework}
\label{sec:method-pipeline}

\noindent\textbf{Coverage anchors.} RiPPLE fixes an anchor set $\cS\subset\cA$ with $|\cS|=n$ once, before
any training, by farthest-point sampling (FPS) in the encoding space. From a seed, each new anchor is the
architecture farthest from those already chosen,
\begin{equation}\label{eq:fps}
\begin{aligned}
  a_{k+1} &= \argmax_{a\in\cA}\ \min_{a'\in\cS_k}\ \lVert \bphi(a)-\bphi(a')\rVert_2,\\
  \cS_{k+1} &= \cS_k\cup\{a_{k+1}\}.
\end{aligned}
\end{equation}
The order is nested, so one list supplies every budget $n$ by truncation, and it inherits a
$k$-center covering guarantee,
\begin{equation}\label{eq:coverage}
  r(\cS_n) \le 2\,r^{\star}_n,
\end{equation}
where $r(\cS_n)$ is the radius that covers $\cA$ with balls at the anchors and $r^{\star}_n$ is the
optimal covering radius over all $n$-subsets~\cite{gonzalez1985clustering}. Coverage spreads the
anchors to fill the space rather than follow its density, so the propagator interpolates inside a
covered support instead of extrapolating into empty regions.

\noindent\textbf{Prefix training and curve reading.} Each anchor $a\in\cS$ is trained for the prefix of
$P=\lceil\pi T\rceil$ epochs, which yields a train-loss curve
$\ell_a=(\ell_a^{(1)},\dots,\ell_a^{(P)})$. A reader $r$ maps this partial curve to a scalar surrogate
label
\begin{equation}\label{eq:yhat}
  \wh{y}_a = r(\ell_a) =
  \begin{cases}
    -\operatorname{mean}\bigl(\ell_a^{(P-k+1)},\dots,\ell_a^{(P)}\bigr), & \text{saturating,}\\[2pt]
    -\operatorname{EMA}_{\gamma}(\ell_a),                                 & \text{noisy,}\\[2pt]
    -\,\widehat{\ell}_a(H),                                              & \text{clean,}
  \end{cases}
\end{equation}
whose branches read the level of the last $k$ epochs, an exponential moving average of the loss, and a
log-linear extrapolation of the curve to a finite horizon $H$~\cite{domhan2015speeding}. The leading
minus sign orients $\wh{y}_a$ so that larger is better. The branch is keyed to the schedule shape and
the horizon $H$ is fixed per space, both decidable before any anchor is trained. We read the training
loss rather than a validation metric, which removes the need for a held-out split at every anchor, and
we write $\wh{y}_a$ rather than $y(a)$ to stress that the label is a prediction from the partial curve,
not a measurement of true performance.

\noindent\textbf{Label propagation.} A propagator $g$ is fit on the anchor pairs
$\{(\bx(a),\wh{y}_a):a\in\cS\}$ and applied to every architecture, giving the score
\begin{equation}\label{eq:score}
  s(a) = g\bigl(\bx(a)\bigr), \qquad \forall\, a\in\cA .
\end{equation}
Spreading the anchor labels to the untrained majority over the label-free feature graph is a
transductive, graph-based label-propagation problem~\cite{zhu2003labelprop}; we use extremely
randomized trees, whose averaging over random splits suits the noisy extrapolated labels. Scoring the
untrained $N-n$ architectures is a single inference pass over precomputed features. The design places
the partial-training signal in the label, not in a feature. A curve used as a per-candidate feature
must be computed for every architecture the predictor scores, and computing it means partially
training that architecture, so ranking the whole space this way would require training all of it.
This is the failure to scale that OMNI~\cite{white2021powerful} reports; placing the signal in the label
instead confines all training to the $n$ anchors, and the untrained majority is scored from
label-free features alone.

\noindent\textbf{Ranking and selection.} The deliverable is the ranking of $\cA$ by $s$, obtained at a
budget of $\fte=n\,\pi$; read as a selector, RiPPLE returns $\argmax_{a\in\cA} s(a)$. The query cost is
negligible, since scoring is a lookup over precomputed features, so the only training cost is the prefix
training of the $n$ anchors. Algorithm~\ref{alg:ripple} summarizes the pipeline.

\begin{algorithm}[t]
\caption{RiPPLE}
\label{alg:ripple}
\begin{algorithmic}[1]
\Require space $\cA$; label-free features $\bx(a)=[\bz(a);\bphi(a)]$, $\forall a\in\cA$; budget $(n,\pi)$; schedule length $T$
\Ensure ranking of $\cA$ by $s$, and selection $a^{\star}=\argmax_{a} s(a)$
\Statex \textit{// coverage anchors (once, no training)}
\State $\cS \gets$ FPS of $n$ anchors \hfill Eq.~\eqref{eq:fps}
\Statex \textit{// prefix training and curve reading (the only training)}
\For{each anchor $a \in \cS$}
  \State train $a$ for $P=\lceil\pi T\rceil$ epochs $\to$ curve $\ell_a$
  \State $\wh{y}_a \gets r(\ell_a)$ \hfill reader Eq.~\eqref{eq:yhat}; no truth
\EndFor
\Statex \textit{// label propagation (inference only)}
\State fit propagator $g$ on $\{(\bx(a),\wh{y}_a): a\in\cS\}$
\State $s(a) \gets g(\bx(a))$, $\forall a\in\cA$
\State \Return $\cA$ ranked by $s$
\end{algorithmic}
\end{algorithm}

Every per-space choice is keyed to a property decidable before any training: the proxy bank and
encodings to the architecture type, and the reader branch to the schedule shape. RiPPLE thus runs one
unchanged pipeline on every space, with no per-benchmark hyperparameter search, and an ablation shows
that this label-free configuration matches a label-reading oracle that is allowed to pick the best
proxy set, encoding, and reader per cell.

\section{Experiments}
\label{sec:exp}

\begin{table}[t]
\centering
\caption{Whole-space ranking across all twelve cells and four search-space families under a common ceiling
of $150$ full-training equivalents (held-out Kendall $\tau$). The graph and sequence predictors are
undefined on the size, recurrent, and transfer families ($-$), and the Weisfeiler--Leman process,
though defined everywhere, collapses on the recurrent cell. RiPPLE-full and RiPPLE-prefix are the only
methods that rank well on every cell; RiPPLE-prefix beats every external baseline on eleven of the
twelve cells (GP-WL edges it by $0.001$ on NATS-SSS IN16, where RiPPLE-full is best). Best per row in
bold.}
\label{tab:breadth}
\small
\setlength{\tabcolsep}{4pt}
\begin{tabular*}{\textwidth}{@{\extracolsep{\fill}}l l ccc cc@{}}
\toprule
Family & Cell & GCN~\cite{dudziak2020brpnas} & SemiNAS~\cite{luo2020seminas} & GP-WL~\cite{ru2021nasbowl} & RiPPLE-full & RiPPLE-prefix \\
\midrule
\multirow{4}{*}{Topology}
 & NAS-Bench-201 C10   & 0.473 & 0.592 & 0.602 & 0.832 & \textbf{0.851} \\
 & NAS-Bench-201 C100  & 0.586 & 0.618 & 0.666 & 0.835 & \textbf{0.846} \\
 & NAS-Bench-201 IN16  & 0.626 & 0.596 & 0.677 & 0.814 & \textbf{0.851} \\
 & NAS-Bench-101 C10   & 0.428 & 0.614 & 0.425 & \textbf{0.682} & 0.675 \\
\midrule
\multirow{3}{*}{Size}
 & NATS-SSS C10        & $-$ & $-$ & 0.824 & 0.841 & \textbf{0.843} \\
 & NATS-SSS C100       & $-$ & $-$ & 0.886 & 0.885 & \textbf{0.891} \\
 & NATS-SSS IN16       & $-$ & $-$ & 0.890 & \textbf{0.892} & 0.889 \\
\midrule
Recurrent
 & NAS-Bench-NLP PTB   & $-$ & $-$ & 0.180 & 0.433 & \textbf{0.509} \\
\midrule
\multirow{4}{*}{Transfer}
 & TransNAS Object     & $-$ & $-$ & 0.498 & 0.666 & \textbf{0.673} \\
 & TransNAS Scene      & $-$ & $-$ & 0.604 & \textbf{0.771} & 0.734 \\
 & TransNAS Jigsaw     & $-$ & $-$ & 0.509 & 0.597 & \textbf{0.672} \\
 & TransNAS AutoEnc    & $-$ & $-$ & 0.576 & \textbf{0.668} & 0.649 \\
\bottomrule
\end{tabular*}
\end{table}

\subsection{Experimental settings}
\label{sec:exp-setup}

\noindent\textbf{Search spaces.} We evaluate on twelve cells drawn from four search-space families that
stress a predictor along different axes. The \emph{topological} family varies the wiring and operations
of a cell: NAS-Bench-201~\cite{dong2020nasbench201} on CIFAR-10, CIFAR-100, and ImageNet16-120, and the
larger NAS-Bench-101~\cite{ying2019nasbench101} on CIFAR-10. The \emph{size} family, NATS-Bench-SSS
~\cite{dong2021natsbench} on the same three datasets, fixes the topology and varies only the channel
widths, so structural graph features are uninformative. The \emph{recurrent} family,
NAS-Bench-NLP~\cite{klyuchnikov2022nasbenchnlp} on Penn Treebank, replaces the convolutional cell with a
recurrent one. The \emph{transfer} family, the four TransNAS-Bench-101~\cite{duan2021transnasbench}
tasks (object and scene classification, jigsaw, and autoencoding), changes the task rather than the
space, including two self-supervised objectives with no classification loss. This spread is deliberate:
a predictor that assumes a labelled classification cell with an informative graph works on the first
family and degrades or is undefined on the others. Beyond the tabular cells we search the DARTS
space~\cite{liu2019darts} of about $10^{18}$ architectures as a selector.

\noindent\textbf{Metrics.} We measure ranking quality as Kendall's $\tau$ and Spearman's $\rho$ between
the predicted scores and the true accuracies. Crucially, we compute them on the strict held-out set
$\cA\setminus\cS$ that excludes every architecture whose curve was read, so no trained anchor inflates
the correlation; this is stricter than the whole-benchmark evaluation the predictor literature reports,
and we align to that convention only in the head-to-head below. Read as a search method, we also report
the test accuracy of the architecture a method selects against the space optimum.

\noindent\textbf{Protocol and baselines.} The main comparison uses a common ceiling of $150$ full-training
equivalents (FTE, one FTE is one architecture trained to completion) on the same coverage anchors, and
is scored on the same strict held-out split with no validation labels beyond the budget. The anchors are
the nested farthest-point order over the architecture encoding, so the budget-$m$ set is the first $m$
of a single deterministic order and every method sees the same coverage. We compare against three
whole-space predictors that, like RiPPLE, consume labels and score the rest of the space: a
graph-convolutional predictor (GCN)~\cite{dudziak2020brpnas}, a Weisfeiler--Leman Gaussian process
(GP-WL)~\cite{ru2021nasbowl}, and a self-training predictor (SemiNAS)~\cite{luo2020seminas}, each the
mean over five seeds. The heavier encoding predictors report only under their own protocols, so we
compare to them both on a label-efficiency frontier and at their nominal sample budgets.
RiPPLE's propagator is a set of extremely randomized trees over label-free features; the full per-space
configuration is deferred to the supplementary.

\subsection{Results}
\label{sec:exp-results}

\noindent\textbf{Whole-space ranking across families.}
\label{sec:exp-breadth}
Table~\ref{tab:breadth} reports held-out ranking quality on all twelve cells, and the story is coverage
as much as accuracy. The graph and sequence predictors are \emph{undefined} on eight of the twelve
cells: GCN and SemiNAS read a cell graph and a classification loss, neither of which the size, recurrent,
or transfer families provide, so they run only on the four topological cells. GP-WL is defined
everywhere through a kernel fallback, but on the recurrent cell its Weisfeiler--Leman features degenerate
and its correlation collapses to $0.180$, against RiPPLE-prefix's $0.509$. On the four topological cells
where every method runs, the gap is still wide: RiPPLE-prefix reaches a Kendall $\tau$ near $0.85$ on the
three NAS-Bench-201 cells while the strongest supervised baseline ranges from $0.60$ to $0.68$ across the three cells. On the size
family GP-WL is genuinely competitive, since channel widths are easy to rank, yet RiPPLE-prefix still
edges it on two of three cells; on the transfer family, including the two label-free reconstruction
tasks, only RiPPLE-prefix and RiPPLE-full produce a usable ranking. Two comparisons inside the table
trade off the label choice against the anchor count: RiPPLE-full spends the whole budget on $150$ fully
trained true labels and is the natural upper control, while RiPPLE-prefix trains $n$ prefix-anchors
($n=167$--$667$, prefix fraction $\pi=0.12$--$0.46$) and reads an extrapolated label, spending about
$43$--$150$ FTE. Across the twelve cells the two have similar correlations, and RiPPLE-prefix is the
higher of the two on eight cells; within the same budget ceiling the cheaper prefix label buys more anchors, and this
extra coverage offsets the noisier label, whereas at matched anchor count the true label is strictly
better (Table~\ref{tab:predlane}, $0.906$ against $0.856$ at $1563$ anchors), so RiPPLE-full is the
genuine upper control and the extra-anchor advantage is a property of the cheap-label design rather than
of label quality. Against the external baselines RiPPLE-prefix is best on eleven of the twelve cells,
edged only by GP-WL on NATS-SSS ImageNet16-120. The margins elsewhere are wide: on the three
NAS-Bench-201 cells RiPPLE-prefix leads the strongest supervised baseline by $0.17$ to $0.25$ in Kendall
$\tau$, and on the recurrent cell it triples the Weisfeiler--Leman correlation from $0.180$ to $0.509$. No reweighting of the same $150$ labels used as
per-candidate features closes these gaps; the gain comes from spending the labels on coverage anchors and
propagating them, the central choice of this work, which is also what makes the pipeline defined at all
on the eight non-topological cells where the graph and sequence predictors cannot run.

\noindent\textbf{Label efficiency.}
\label{sec:exp-frontier}
\begin{table}[t]
\centering
\caption{First reported budget reaching Kendall $\tau\geq0.85$ on NAS-Bench-201
(CIFAR-10). RiPPLE's $127$ FTE is the first crossing in its tested grid, not a proven
continuous minimum. External results use their respective source protocols, not a
controlled common-cost evaluation.}
\label{tab:frontier}
\small
\begin{threeparttable}
\begin{tabular*}{0.78\linewidth}{@{\extracolsep{\fill}}lr@{}}
\toprule
Method & Labels to $\tau\!=\!0.85$ \\
\midrule
\textbf{RiPPLE-prefix (ours)}\tnote{b}        & \textbf{127} \\
\midrule
NN-Former~\cite{xu2025nnformer}               & $469$  \\
NAR-Former~V2~\cite{yi2023narformerv2}        & $781$  \\
FlowerFormer~\cite{hwang2024flowerformer}\tnote{c} & $\approx391+40$ \\
TA-GATES~\cite{ning2022tagates}\tnote{c}       & $\approx391+40$ \\
NAR-Former~\cite{yi2023narformer}             & $1563$ \\
TNASP~\cite{lu2021tnasp}                      & $>\!1563$ \\
Five further predictors\tnote{a}             & $>\!1563$ \\
\bottomrule
\end{tabular*}
\begin{tablenotes}\footnotesize
\item[a] Neural Predictor, Graphormer, NAO, GraphTrans, and PINAT do not reach $0.85$
within the largest budget they report. Sources: the uniform NAS-Bench-201 tables of
NN-Former~\cite{xu2025nnformer} and FlowerFormer~\cite{hwang2024flowerformer}.
\item[b] The $127$ FTE counts anchor training; the one-time proxy-bank feature cost is
reported separately in Table~\ref{tab:cost}. No uniform total-cost ratio is inferred.
\item[c] The FlowerFormer table reports $5\%$ of a $7{,}813$-architecture training pool
(approximately $391$ training labels), plus $40$ validation labels and an independent
test set. Integer rounding follows the source implementation. Other predictor rows
are sourced from the NN-Former compilation; their validation costs are not added here.
\end{tablenotes}
\end{threeparttable}
\end{table}

Read at a fixed quality bar, RiPPLE reaches a target ranking with far fewer labels than the strongest
encoding predictors. Table~\ref{tab:frontier} lists the budget each method needs to reach a Kendall
$\tau$ of $0.85$ on NAS-Bench-201 (CIFAR-10), using protocol-specific reported results.
RiPPLE-prefix first reaches this quality at the tested point of $127$ full-training equivalents;
the external predictor entries retain their distinct training, validation, and evaluation conventions.
Figure~\ref{fig:teaser} presents a budget-quality plot as a descriptive comparison,
not a strictly matched continuous cost frontier. The efficiency is a
direct consequence of the label-not-feature design: because the budget is spent training a few anchors
rather than labelling many candidates, the cost scales with the anchor count and not with the space, so
RiPPLE reaches a strong ranking from a small anchor-training budget without implying a
controlled cost ratio against predictors evaluated under other protocols.

\noindent\textbf{Head-to-head under the predictor protocol.}
\label{sec:exp-predlane}
\begin{table}[t]
\centering
\caption{Head-to-head under the predictor papers' own protocol on NAS-Bench-201: train on a random
subset with true labels, test on the whole $15{,}625$-architecture space, report Kendall's $\tau$ at
the standard budgets. On the full space, at matched supervision, RiPPLE-full is the strongest predictor
in every budget column; the prefix variant, at a fraction of the training cost, already exceeds the
best prior method at the smallest budget. RiPPLE augments the standard encoding with a zero-cost proxy
bank the transformer baselines do not use, so this comparison is our proxy-augmented feature set with a
tree propagator against their learned representations, not the label mechanism. Best per column in bold;
full roster and NAS-Bench-101 in the supplementary.}
\label{tab:predlane}
\small
\setlength{\aboverulesep}{0pt}
\setlength{\belowrulesep}{0pt}
\renewcommand{\arraystretch}{1.15}
\begin{tabular}{@{}lcccc}
\specialrule{\heavyrulewidth}{0pt}{2pt}
Method & \makecell{$1\%$\\(156)} & \makecell{$3\%$\\(469)} & \makecell{$5\%$\\(781)} & \makecell{$10\%$\\(1563)} \\
\specialrule{\lightrulewidth}{2pt}{0pt}
TNASP~\cite{lu2021tnasp}                  & 0.589 & 0.640 & 0.689 & 0.724 \\
NAR-Former~\cite{yi2023narformer}         & 0.660 & 0.790 & 0.849 & \cs 0.901 \\
PINAT~\cite{lu2023pinat}                  & 0.631 & 0.706 & 0.761 & 0.784 \\
NAR-Former V2~\cite{yi2023narformerv2}    & 0.752 & 0.846 & 0.874 & 0.888 \\
NN-Former~\cite{xu2025nnformer}           & 0.804 & \cs 0.860 & \cs 0.879 & 0.890 \\
\midrule
\textbf{RiPPLE-full} (ours)               & \cb\textbf{0.832} & \cb\textbf{0.870} & \cb\textbf{0.887} & \cb\textbf{0.906} \\
\textbf{RiPPLE-prefix} (ours)            & \cs 0.821 & 0.844 & 0.851 & 0.856 \\
\bottomrule
\end{tabular}
\end{table}

To remove any doubt from reading budgets off other papers, we run RiPPLE inside the predictor protocol
itself, standardized by TNASP~\cite{lu2021tnasp} and reused unchanged by NAR-Former~\cite{yi2023narformer},
PINAT~\cite{lu2023pinat}, NAR-Former~V2~\cite{yi2023narformerv2}, and NN-Former~\cite{xu2025nnformer}:
train the predictor on a random subset with true labels, then evaluate on the whole space by Kendall's
$\tau$. We use these nominal sample budgets on NAS-Bench-201, whose $15{,}625$-architecture space we hold, at
the four standard budgets (Table~\ref{tab:predlane}). At matched supervision RiPPLE-full is the strongest
predictor in every budget column, above the two most recent transformers, doing so with a CPU tree
ensemble over label-free features that pair zero-cost proxies with graph encodings, a feature set the
transformers do not receive. RiPPLE-prefix is more striking on cost: spending about $30$ full-training
equivalents against the $156$ full trainings of the $1\%$ column, it already exceeds the strongest prior
method there, and it saturates near $0.86$, the ceiling set by the agreement between the early signal and
the final test accuracy. The margin is consistent rather than a single lucky column: RiPPLE-full leads
NN-Former by $0.028$, $0.010$, $0.008$, and $0.016$ at the four budgets, and leads NAR-Former~V2 by
$0.01$ or more throughout, so a training-label-free feature set fed to a tree propagator matches or leads
the trained graph transformers on this benchmark. The proxy bank drives that margin, since the
encoding-only variant falls below both transformers (supplementary); what the head-to-head shows is that
this proxy-augmented feature set is strong, and the prefix result shows the label that trains it can be
cheap without giving that strength up. The full roster and the NAS-Bench-101 comparison, where the deployed feature set cannot
cover the whole $423{,}624$-architecture space and RiPPLE is mid-tier among the learned graph
representations, are in the supplementary.

\noindent\textbf{Selection.}
\label{sec:exp-selection}
\begin{table}[t]
\centering
\caption{Selection quality: test accuracy (\%) of the best of each method's top-5 predicted
architectures, against the held-out evaluation-pool optimum, under a $150$-FTE ceiling. RiPPLE-prefix selects the
strongest architecture on the three NAS-Bench-201 datasets and lands within a fraction of a point of
the optimum; on NAS-Bench-101 the methods tie within noise. Best per column (excluding the oracle)
shaded dark, second-best light.}
\label{tab:selection}
\small
\setlength{\tabcolsep}{5pt}
\setlength{\aboverulesep}{0pt}
\setlength{\belowrulesep}{0pt}
\renewcommand{\arraystretch}{1.15}
\begin{tabular}{@{}lcccc}
\specialrule{\heavyrulewidth}{0pt}{2pt}
Method & \makecell{NB201\\C10} & \makecell{NB201\\C100} & \makecell{NB201\\IN16} & \makecell{NB101\\C10} \\
\specialrule{\lightrulewidth}{2pt}{0pt}
GCN~\cite{dudziak2020brpnas}   & 93.46 & 71.13 & 44.17 & 93.38 \\
SemiNAS~\cite{luo2020seminas}  & 93.44 & \cs 71.24 & \cs 46.04 & \cb\textbf{94.31} \\
GP-WL~\cite{ru2021nasbowl}     & 92.90 & 70.98 & 45.88 & 93.48 \\
RiPPLE-full                    & \cs 93.47 & 70.77 & 45.91 & 94.15 \\
\textbf{RiPPLE-prefix}         & \cb\textbf{94.29} & \cb\textbf{72.94} & \cb\textbf{46.48} & \cs 94.27 \\
\midrule
Oracle (held-out pool)         & 94.37 & 73.20 & 47.31 & 94.54 \\
\bottomrule
\end{tabular}
\end{table}

A ranking is useful only if its top is trustworthy, so we read the predictor as a selection rule and
report the test accuracy of the best of each method's top-five architectures against the held-out pool optimum
(Table~\ref{tab:selection}). RiPPLE-prefix selects the strongest architecture among the equal-budget
methods on the three NAS-Bench-201 datasets and lands within $0.08$ to $0.83$ points of the optimum; on
NAS-Bench-101 the methods agree within noise. The margin is small in absolute accuracy because the top of
these spaces is flat, which is exactly why a predictor with high global rank quality but only modest
top-tier discrimination still selects a near-optimal network. Under the standard zero-cost protocol that
scores $3000$ sampled candidates and deploys the best, RiPPLE-prefix selects the strongest architecture
among thirteen methods on all three NAS-Bench-201 datasets (supplementary).

\noindent\textbf{Scaling to a large space.}
\label{sec:exp-scale}
The same pipeline runs unchanged on the DARTS space of about $10^{18}$ architectures, with ground truth
from NAS-Bench-301 and no retuning. Ground truth here is the NAS-Bench-301 surrogate rather than measured
training, so the reported correlations are bounded by that surrogate's fidelity and should be read
relative to the baselines rather than as absolute rank agreement with true accuracy; the $2.61\%$
selection result below is measured by real training and does not depend on the surrogate. RiPPLE ranks above the fidelity baselines at every budget, and the
extrapolated label costs about $0.06$ in correlation against the true one, the same small gap seen on the
tabular cells. Its selected architecture, fully trained under the standard DARTS protocol over five
seeds, reaches a test error of $2.61\%$, within the band of recent differentiable search. We do not claim
to beat a dedicated search on the single architecture it returns: run to the same budget, multi-fidelity
search matches RiPPLE on best-found accuracy to within one point. What RiPPLE adds is a ranking of the
entire space at a fractional budget, read out at any point without rerunning. The DARTS result is the
strongest test of the central claim: the space is fourteen orders of magnitude larger than the tabular
cells, yet the same anchors-plus-propagation pipeline, with no new hyperparameters, ranks it well and the
extrapolated label loses only $0.06$, exactly as on the small benchmarks.

\needspace{4\baselineskip}%
\subsection{Ablation study}
\label{sec:exp-ablation}
\begin{figure}[t]
\centering
\begin{minipage}[c]{0.455\textwidth}
\includegraphics[width=\linewidth]{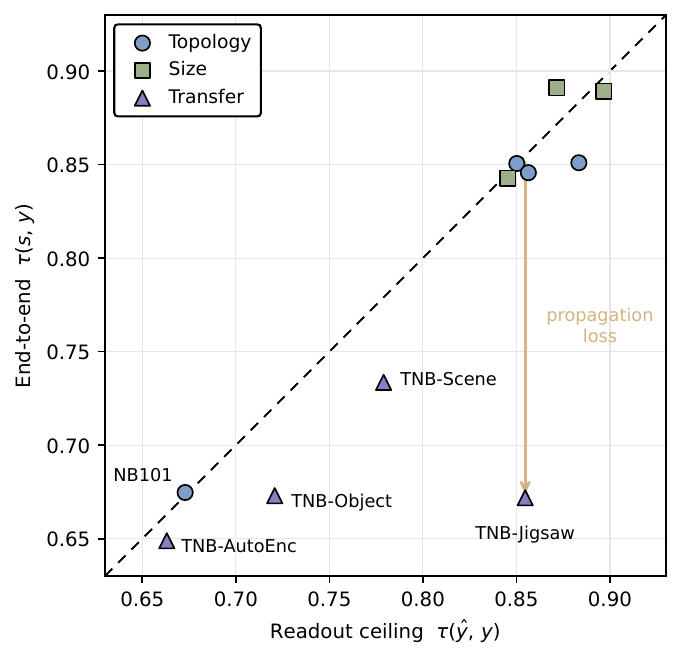}
\end{minipage}\hfill
\begin{minipage}[c]{0.51\textwidth}
\caption{Descriptive diagnostic of the readout and propagation stages, one point per cell.
The horizontal axis is the component readout correlation $\tau(\wh{y},y)$ and the vertical
axis is the end-to-end held-out correlation $\tau(s,y)$. Points are coloured by
search-space family; the recurrent cell is omitted. These component evaluations
provide a qualitative diagnostic rather than an exact decomposition of each
end-to-end run: proximity to the diagonal alone does not establish a quantitative
propagation loss.}
\label{fig:decomp}
\end{minipage}
\end{figure}

\begin{wraptable}{r}{0.50\textwidth}
\vspace{-1.2\baselineskip}
\centering
\caption{Each stage is necessary. Held-out Spearman $\rho$ averaged over the twelve cells. Spending the
budget on anchor labels rather than aggregating proxies, and separating the read-out from the
propagation rather than fitting a single joint model, each help. Best in bold.}
\label{tab:stageabl}
\small
\begin{tabular}{@{}lc@{}}
\toprule
Variant & Mean $\rho$ \\
\midrule
Proxy aggregation, no anchor labels          & 0.593 \\
Joint feature--epoch model, single stage     & 0.887 \\
\textbf{RiPPLE-prefix}, read then propagate  & \textbf{0.902} \\
\bottomrule
\end{tabular}
\vspace{-0.8\baselineskip}
\end{wraptable}
We ablate the pipeline along three axes: the two-stage design, the feature blocks, and the label-free
rules. \textbf{(1) Each stage is necessary} (Table~\ref{tab:stageabl}). Aggregating the same proxies with
no anchors reaches a mean held-out correlation of only $0.593$ against RiPPLE-prefix's $0.902$, so the anchors
carry real signal; reading the anchor curves without propagating ranks the anchors well but covers at
most $16\%$ of a space, so propagation is what extends the ranking to the untrained majority; and
regressing on features and epoch jointly, rather than reading a label and then propagating it, trails the
two-stage separation. Figure~\ref{fig:decomp} compares the readout and end-to-end component
correlations descriptively; it is not a verified same-run decomposition of the observed
loss. \textbf{(2) Both
feature blocks matter, and which dominates depends on the family} (Table~\ref{tab:featabl}). With the
anchors and reader fixed, the zero-cost proxies lead on the topological cells while the architecture
encoding leads on the size cells, so concatenating both is best on eleven of twelve cells and is edged by
the encoding on the twelfth. This is why RiPPLE uses both blocks and lets the propagator down-weight the rest
rather than selecting a per-space subset. \textbf{(3) The label-free rules need no per-benchmark tuning.}
The feature, reader, and encoding rules of Section~\ref{sec:method-pipeline} read no true label, yet they
recover what a per-cell oracle with access to the true labels would choose, to a mean correlation of
$0.902$; a minimal universal feature set trails this by up to $0.022$ and a single fixed encoding by
$0.015$, so the transfer really does come from the rules rather than from tuning. Together the three
findings validate the design as a whole: the two-stage separation is what turns a partial-training signal
into a whole-space ranking, the label-free feature and reader rules are what let one pipeline transfer
across four families without a single per-benchmark hyperparameter, and neither reads a true label at
deployment. The propagator and reader sweeps, the encoding study, the coverage certificate, and the
tier-stratified precision are reported in full in the supplementary.
\begin{table}[t]
\centering
\caption{Both feature blocks matter, and which one dominates depends on the family. Held-out Kendall
$\tau$ with the anchors and reader fixed, toggling the zero-cost proxies $\bz$ and the architecture
encoding $\bphi$ (a checkmark means the block is used). The proxies lead on the topological cells and the
encoding leads on the size cells, so concatenating both is best on eleven of twelve cells and is edged by
the encoding on the twelfth; this is why RiPPLE uses both blocks and lets the propagator down-weight the
rest. Best per column in bold.}
\label{tab:featabl}
\footnotesize
\setlength{\tabcolsep}{4pt}
\begin{tabular*}{\textwidth}{@{\extracolsep{\fill}}cc cccc ccc c cccc@{}}
\toprule
& & \multicolumn{4}{c}{Topology} & \multicolumn{3}{c}{Size} & Rec. & \multicolumn{4}{c}{Transfer} \\
\cmidrule(lr){3-6}\cmidrule(lr){7-9}\cmidrule(lr){10-10}\cmidrule(lr){11-14}
$\bz$ & $\bphi$ & C10 & C100 & I16 & 101 & C10 & C100 & I16 & NLP & Obj & Scn & Jig & AE \\
\midrule
\yes &      & 0.846 & 0.833 & 0.842 & 0.639 & 0.812 & 0.774 & 0.851 & 0.438 & 0.672 & 0.726 & 0.645 & 0.532 \\
     & \yes & 0.714 & 0.710 & 0.714 & 0.634 & 0.842 & \textbf{0.893} & 0.880 & 0.411 & 0.656 & 0.713 & 0.663 & 0.648 \\
\yes & \yes & \textbf{0.851} & \textbf{0.845} & \textbf{0.851} & \textbf{0.676} & \textbf{0.843} & 0.891 & \textbf{0.889} & \textbf{0.509} & \textbf{0.675} & \textbf{0.734} & \textbf{0.673} & \textbf{0.650} \\
\bottomrule
\end{tabular*}
\end{table}

\section{Conclusion}
\label{sec:conclusion}

We presented RiPPLE, a performance predictor that ranks an entire architecture space at a fractional
training budget by spending the partial-training signal as a label on a few coverage anchors rather
than as a feature on every candidate. This single choice decouples the training cost, which scales
with the anchors, from the ranking reach, which spans the whole space, and it is what lets one
unchanged pipeline rank the tabular benchmarks and the DARTS space of about $10^{18}$ architectures
without retuning. Across twelve cells from four search-space families RiPPLE is the strongest ranker
under a common budget ceiling, it is competitive with its own full-training control while paying a fraction of the label
cost, and its top-ranked networks select architectures within a fraction of a point of the
held-out evaluation-pool optimum on NAS-Bench-201 CIFAR-10 and CIFAR-100.

\paragraph{Limitations.} RiPPLE ranks the whole space, but it does not beat a dedicated multi-fidelity
search on the single architecture that search returns; the two solve different problems and achieve
comparable best-found accuracy under the reported protocols, without establishing statistical equivalence. The curve reader assumes a schedule it can extrapolate and retreats to a
non-extrapolating estimator when that assumption fails, so schedules that neither saturate nor decay
cleanly are read conservatively. Our coverage analysis bounds a conservative nearest-neighbor
propagator rather than the deployed regressor, which ranks strictly better; closing this gap, together
with an adaptive choice of anchors and a learned curve reader, is left to future work.

{
    \small
    \bibliographystyle{ieeenat_fullname}
    \bibliography{main}
}

\clearpage
\maketitlesupplementary
\appendix
\renewcommand{\topfraction}{0.95}
\renewcommand{\textfraction}{0.05}
\renewcommand{\floatpagefraction}{0.85}
\setcounter{topnumber}{3}
\setcounter{totalnumber}{5}
\makeatletter
\setlength{\@fptop}{0pt}
\setlength{\@fpsep}{18pt}
\setlength{\@fpbot}{0pt plus 1fil}
\makeatother

This supplement collects material deferred from the main paper: the deployed per-space configuration
(App.~\ref{sec:sup-config}), the theoretical analysis (App.~\ref{sec:sup-theory}), the cost and
FTE-fairness measurements (App.~\ref{sec:sup-cost}), the label-free selection of the reader horizon
(App.~\ref{sec:sup-backcast}), extended ablations (App.~\ref{sec:sup-abl}), selection and scaling
results (App.~\ref{sec:sup-scale}), and a comparison of RiPPLE under the predictor-lane papers' own
protocols (App.~\ref{sec:sup-protocol}).

\section{Deployed configuration}
\label{sec:sup-config}
CoRA-NAS~\cite{yang2026cora} forms a coarse zero-cost ranking prior and refines it by
propagating anchor residuals derived from early learning curves. RiPPLE instead
propagates prefix-derived surrogate labels directly from coverage-selected anchors,
with zero-cost proxies serving as input features rather than a separately corrected prior.

The pipeline is fixed before any benchmark is seen. The propagator is a set of extremely randomized
trees with $400$ estimators, a maximum feature fraction of $0.5$, a minimum of two samples per leaf,
and no bootstrap, used unchanged on every cell. The zero-cost proxy bank comprises the standard
suite (synflow, snip, grasp, fisher, grad-norm, jacob-cov, NASWOT, Zen, l2-norm, plain, params,
flops) together with ZiCo, MeCo, and SWAP, with the verified per-family banks listed in
Table~\ref{tab:app-zcpzoo}; on the size cells, whose architectures differ only in
channel width, the compact capacity subset is used. The architecture encoding concatenates every
candidate encoding available for a space (operation and adjacency one-hot, path, GRAF where
defined with GRAF-lite on NAS-Bench-101, and width codes on the size space), and duplicate coordinates are dropped. The reader
branch is keyed to the schedule shape: a long clean schedule is extrapolated to a finite horizon, a
noisy one is read by an exponential moving average of the loss, and an early-saturating or short one
is read at its level. Table~\ref{tab:family-rules} states the three per-space rules, each keyed to a
property decidable before any training, so the choice transfers to a new space, and Table~\ref{tab:the-table} lists the deployed instance for
every cell: the anchor count $n$, the prefix epochs trained per anchor, and the resulting budget in full-training equivalents ($\fte=n\pi$, with $\pi$ the prefix fraction of the schedule). The deployed budgets span $43$ FTE on SSS-C10 to $490$ FTE on NAS-Bench-101; ten of the twelve cells use $667$ anchors, with SSS-C10 at $167$ and NAS-Bench-101 at $1058$.

\begin{table}[t]
\centering
\caption{The three per-space rules of the deployed pipeline. Per-space instances are listed in \autoref{tab:the-table}.}
\label{tab:family-rules}
\begin{tabularx}{\linewidth}{@{}l l X@{}}
\toprule
Component & Keyed to & Rule \\
\midrule
Features $\bz$ & architecture type (available proxies) & use the full available bank, no per-space subset selection (the propagator down-weights the rest); size space ships a compact set for parsimony only \\
Reader $r$ & schedule shape & long clean schedule: extrapolate to a fixed horizon; short or saturating: read the level without extrapolation \\
Encoding $\bphi$ & architecture type (available encodings) & concatenate all available encodings, no per-space selection (the propagator down-weights the rest); matches a label-reading oracle (App.~\ref{sec:sup-abl}) \\
\bottomrule
\end{tabularx}
\end{table}

\begin{table}[t]
\centering
\caption{Deployed per-space configuration of RiPPLE for all twelve cells. Features, reader, and encoding follow the rules of \autoref{tab:family-rules}.}
\label{tab:the-table}
\small
\begin{tabular*}{\linewidth}{@{\extracolsep{\fill}}llllcr@{}}
\toprule
Space & Features & Encoding & Reader & Horizon & Anchors$\times$prefix ($\fte$) \\
\midrule
NB201-C10   & full           & GRAF              & exp-decay & $100$ & $667\times38$ ($127$) \\
NB201-C100  & full           & GRAF              & exp-decay & $100$ & $667\times50$ ($167$) \\
NB201-IN16  & full           & GRAF              & exp-decay & $100$ & $667\times38$ ($127$) \\
\midrule
SSS-C10     & compact        & width one-hot     & level     & ---   & $167\times23$ ($43$)  \\
SSS-C100    & compact        & width one-hot     & level     & ---   & $667\times15$ ($111$) \\
SSS-IN16    & compact        & width one-hot     & level     & ---   & $667\times15$ ($111$) \\
\midrule
NB101       & full           & GRAF-lite         & TSE-EMA   & ---   & $1058\times50$ ($490$) \\
\midrule
NLP         & uni6 + labeled & GRAF              & exp-decay & $16$  & $667\times6$ ($80$)   \\
\midrule
TNB-Object  & full           & adjacency one-hot & exp-decay & $P$   & $667\times6$ ($160$)  \\
TNB-Scene   & full           & adjacency one-hot & exp-decay & $P$   & $667\times4$ ($107$)  \\
TNB-Jigsaw  & gen6           & adjacency one-hot & exp-decay & $P$   & $667\times6$ ($160$)  \\
TNB-AutoEnc & gen6           & adjacency one-hot & exp-decay & $P$   & $667\times6$ ($160$)  \\
\bottomrule
\end{tabular*}
\end{table}

\section{Theoretical analysis}
\label{sec:sup-theory}
We give three accounts of RiPPLE: how the end-to-end ranking decomposes into a readout and a
propagation stage, how a fractional budget should be split between anchors and prefix, and what the
coverage of the anchors guarantees. We are explicit about the boundary between what we prove and what
we measure.

\begin{theorem}[Two-stage decomposition]
\label{thm:sandwich}
Let $\tau(\cdot,\cdot)$ be Kendall's rank correlation on a common set of architectures. On a common
set with a global tie-break under which $\wh{y}$ is tie-free,
\begin{equation}\label{eq:sandwich}
  \tau(s,y) = \tau(s,\wh{y})\,\tau(\wh{y},y) + \Cov\!\bigl(\sigma^{s,\wh{y}},\,\sigma^{\wh{y},y}\bigr),
\end{equation}
where $\sigma^{u,v}$ is the vector of pairwise concordance signs, and the end-to-end correlation is
sandwiched by $\tau(s,\wh{y}) + \tau(\wh{y},y) - 1 \le \tau(s,y) \le 1 - |\tau(s,\wh{y}) -
\tau(\wh{y},y)|$.
\end{theorem}

The identity~\eqref{eq:sandwich} is exact on the stated common set, and its residual is
a covariance of pairwise sign alignments. The product of readout and propagation quality
is the leading term, not an identity without the covariance term. Figure~\ref{fig:decomp}
provides a qualitative component diagnostic rather than a numerical verification of
this identity on paired runs.

\begin{proposition}[Compute-optimal allocation]
\label{prop:allocation}
Model the excess ranking error as $e(n,p)=c_0 + a\,n^{-\alpha} + b\,p^{-\beta}$ under the cost
$\fte=n\,p$. Let $B_{\mathrm{crit}}=(\alpha a/\beta b)^{1/\alpha}$ and
$\gamma=\alpha\beta/(\alpha+\beta)$. For $B<B_{\mathrm{crit}}$ the optimum is interior, with
$n^{*}(B)\propto B^{\beta/(\alpha+\beta)}$ and $p^{*}(B)\propto B^{\alpha/(\alpha+\beta)}$, and both
budget corners are strictly suboptimal; for $B\ge B_{\mathrm{crit}}$ the optimum is the full-training
corner.
\end{proposition}

\begin{figure}[htbp]
\centering
\includegraphics[width=\linewidth]{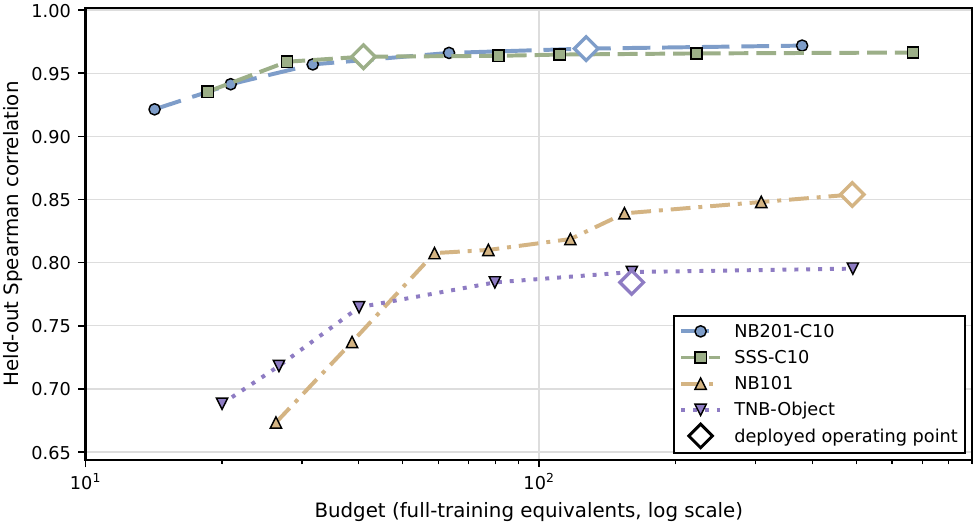}
\caption{Budget-quality frontier for four representative cells (\autoref{prop:allocation}): held-out correlation against budget in full-training equivalents (log scale). Diamonds mark the original allocation-grid operating points. The SSS-C10 point uses $22$ epochs (about $41$ FTE), while the current $23$-epoch deployment costs about $43$ FTE.}
\label{fig:fte-frontier}
\end{figure}

The fractional-training regime RiPPLE occupies is therefore an interior optimum of the budget rather
than a compromise: below $B_{\mathrm{crit}}$ neither training a few anchors fully nor training many
anchors trivially is best. We locate the operating point at the knee of the empirical
budget-quality frontier (Fig.~\ref{fig:fte-frontier}). Held-out correlation rises with the budget and saturates at a knee, where the deployed operating point sits; the steep-frontier spaces reach their knee within tens of full-training equivalents, while the mixed space NAS-Bench-101 needs more.

\begin{proposition}[Prefix law]
\label{prop:prefix}
Under a log-linear curve model with uncorrelated per-epoch noise $\sigma_z$, the standard error of the
horizon label satisfies $\mathrm{SE}(\wh{y})\approx 2\sqrt{3}\,\sigma_z\,p^{-3/2}\,T^{-1/2}$ for
$p\ll1$, so a target error $\varepsilon$ requires $p\ge(12\sigma_z^2/(\varepsilon^2 T))^{1/3}$.
\end{proposition}

\begin{figure}[t]
\centering
\includegraphics[width=\linewidth]{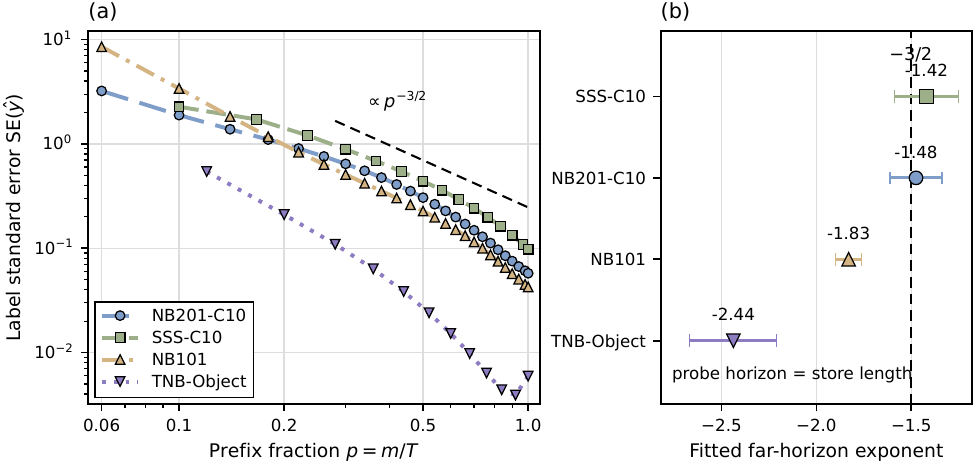}
\caption{Empirical test of the prefix law (\autoref{prop:prefix}). (a) Standard error of the extrapolated label against the prefix fraction $p=m/T$, with the predicted $p^{-3/2}$ reference (dashed); (b) per-cell fitted far-horizon exponents with $95\%$ confidence intervals.}
\label{fig:prefix-law}
\end{figure}

The required prefix falls as the cube root of the noise-to-tolerance ratio and shrinks with the
schedule length, which is why a short prefix suffices on the long clean schedules of the tabular
cells and is marginal on the short schedules of the transfer tasks. The measured label error follows
this law, with a fitted exponent of $-1.575$ across the far-horizon cells
(Fig.~\ref{fig:prefix-law}). The individual far-horizon slopes range from $-1.42$ to $-1.83$ around the predicted $-3/2$. The transfer cell TNB-Object fits at $-2.44$ because its probe horizon equals the store length, so the fit lies outside the far-horizon regime the law assumes, and the steeper slope is an artefact of that boundary rather than a violation of the law.

\begin{lemma}[Coverage certificate]
\label{lem:coverage}
The nested farthest-point order satisfies $r(\cS_n)\le 2\,r^{\star}_n$ for every prefix $n$, with
$r^{\star}_n$ the optimal covering radius. If the target is $L$-Lipschitz in the coverage metric
outside an exceptional mass, the nearest-neighbor propagator with anchor labels $\wh{y}$ obeys, off
that mass, $\sup_x|g_{\mathrm{1NN}}(x)-y(x)|\le 2L\,r^{\star}_n + \max_{s\in\cS_n}|\wh{y}_s-y(s)|$, and
every pair whose true gap exceeds twice this bound is ordered correctly.
\end{lemma}

\begin{figure}[t]
\centering
\includegraphics[width=\linewidth]{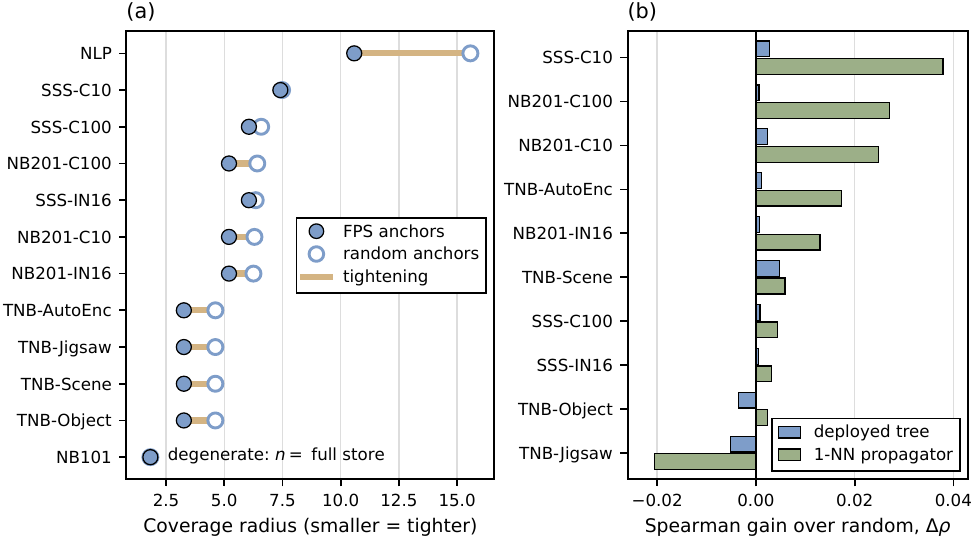}
\caption{Farthest-point against random anchors (\autoref{lem:coverage}). (a) Per-cell coverage radius under each anchor order; (b) downstream Spearman gain over random anchors for the deployed tree and the nearest-neighbor propagator.}
\label{fig:fps-coverage}
\end{figure}

The certificate bounds a conservative nearest-neighbor propagator; the deployed tree reads proxy
coordinates outside the coverage metric and ranks strictly better on every cell, so we treat the gap
between the bound and the deployed tree as a measured gain rather than a claim of the lemma
(Fig.~\ref{fig:fps-coverage}). The farthest-point order attains a smaller coverage radius than random anchors on eleven of twelve cells; NAS-Bench-101 is the exception and is degenerate, since its anchor budget equals the full store and the two orders coincide. Downstream, over the ten non-degenerate cells, the deployed tree with farthest-point anchors is at least as good as with random anchors on eight (mean $\Delta\rho=+0.0005$), and the gain is clearer for the nearest-neighbor rule the certificate actually bounds (mean $\Delta\rho=+0.0115$); the near-degenerate recurrent cell NLP, whose pool barely exceeds the budget, is excluded from these means, and the tree loses to random on the two flat-frontier transfer cells TNB-Object and TNB-Jigsaw.

\section{Cost and FTE fairness}
\label{sec:sup-cost}
We measure, on a stratified sample of $16$--$24$ architectures per space spanning the FLOPs range, the
full zero-cost proxy bank time, one training epoch, and the encoding time on a pinned L40S. The proxy
bank costs a small fraction of one epoch per architecture (median $0.64$~s against $1.75$~s on
NAS-Bench-201, $3.4$~s against $9.6$~s on NAS-Bench-101). Because scoring the whole space needs a
feature vector for every architecture, the feature cost scales with $N$ and is sub-dominant but not
negligible; we report it as GPU-hours per space alongside the anchor-training budget.

FTE counts
training epochs and per-architecture epoch time varies by $28$ to $194\times$ within a space, so FTE
is not a wall-clock proxy on its own. The fairness of the matched-FTE comparison rests on the
distribution of per-architecture cost: RiPPLE's coverage anchors and the sampled baselines have
matching FLOPs distributions, since both draw from the same nested farthest-point order, so $150$ FTE
and $150$ full-training labels cost the same wall-clock in expectation. 

Table~\ref{tab:cost} reports the per-method predictor and feature cost. RiPPLE fits a CPU tree ensemble in $1.6$~s and trains no network on a GPU, unlike the GCN ($10.8$~s) and SemiNAS ($32.7$~s); it is not the cheapest predictor, since the closed-form GP-WL kernel fits in $0.09$~s. What RiPPLE alone pays is the zero-cost proxy bank, about $0.6$~s per architecture, roughly $0.4$ of one training epoch, summing to about $19\%$ of the $150$-FTE anchor budget over the whole space; that feature cost is what lifts its ranking quality far above the kernel and graph predictors.

\begin{table}[t]
\centering
\caption{Per-method predictor and feature cost under the common $150$-FTE ceiling. Fit time is measured on the reported device; feature cost per architecture is measured on a pinned L40S.}
\label{tab:cost}
\footnotesize
\setlength{\tabcolsep}{3.5pt}
\begin{tabular*}{\linewidth}{@{\extracolsep{\fill}}llcl@{}}
\toprule
Method & Predictor & Fit time & Feature / arch \\
\midrule
GCN~\cite{dudziak2020brpnas}   & $300$-epoch GNN        & $10.8$~s (GPU) & encoding, $\sim\!0$ \\
SemiNAS~\cite{luo2020seminas}  & pretrain $+$ finetune  & $32.7$~s (GPU) & encoding, $\sim\!0$ \\
GP-WL~\cite{ru2021nasbowl}     & closed-form kernel     & $0.09$~s (CPU) & encoding, $\sim\!0$ \\
\textbf{RiPPLE} (ours)         & CPU tree ensemble      & $1.6$~s (CPU)  & proxy bank, $0.6$~s \\
\bottomrule
\end{tabular*}
\end{table}

\section{Label-free selection of the reader horizon}
\label{sec:sup-backcast}
The deployed reader uses a fixed per-space horizon. That horizon can be chosen without any label by a
backcast on the observed prefix: fit the extrapolator on the first $60\%$ of the prefix, predict the
last observed epochs, and select the horizon multiplier that best reproduces them, reading no true
accuracy. This label-free choice recovers the horizon that a true-label oracle would pick on nine of
twelve cells, which is why the reader carries no per-benchmark tuning.

\begin{figure}[t]
\centering
\includegraphics[width=\linewidth]{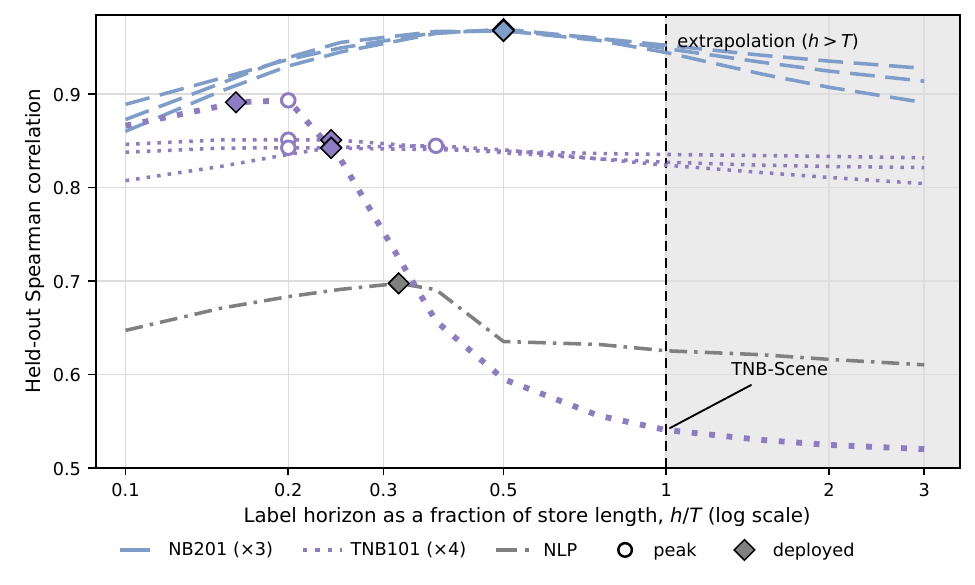}
\caption{Held-out Spearman correlation against the reading horizon, expressed as a fraction of the store length $T$. Open circles mark each cell's peak, diamonds the deployed horizon, and shading the region past the store length.}
\label{fig:horizon}
\end{figure}

Fig.~\ref{fig:horizon} shows the sweep this choice operates on. Each exponential-decay cell rises to a strict interior optimum and then declines monotonically, and the deployed horizon sits at or near the peak in every cell; reading the label at or beyond the store length instead of at the peak degrades the ranking, from about $1.5$ points on the NB201 cells up to a $35$-point cliff on TNB-Scene, whose correlation falls from $0.893$ at its peak to $0.541$ at $T$. The interior optimum is what justifies the $\min(\cdot,T)$ clamp: short or early-saturating schedules are read at their level rather than extrapolated into this collapse.

\section{Extended ablations}
\label{sec:sup-abl}
This section reports in full the component ablations summarized in the main paper, beyond the
necessity ablation reported there. The ablations cover five axes in turn: the proxy bank, the propagator, the reading rule, the architecture encoding, and the discrimination at the top of the ranking.

\subsection{Proxy bank}
\label{sec:sup-abl-bank}
Feature selection is unnecessary in the deployed pipeline. The full proxy bank that the family rule
admits for a space, used together with the architecture encoding, matches a per-cell leakage oracle
that reads the true labels to pick the best proxy subset, to within $0.001$ in aggregate and $0.003$
on every cell (Table~\ref{tab:family-oracle}); a minimal universal set forced across all spaces
trails the bank by up to $0.022$ on NAS-Bench-101 and $0.016$ on ImageNet16-120, while on the label-free cells
the two coincide and on the size cells they agree to within $0.001$. The bank itself is fixed by what each space exposes rather
than chosen (Table~\ref{tab:app-zcpzoo}): the CNN topology cells expose all eighteen proxies, the
channel-size space a compact set of seven, since a topology-invariant space makes most structural
probes degenerate, the recurrent space seven recurrent-specific proxies, and the reconstruction and
self-supervised cells only the six universal ones, because saliency and gradient probes such as
SNIP, GraSP, Fisher, and grad-norm assume a classification loss. RiPPLE therefore performs no
per-space selection: it reads whichever proxies a space defines and lets the propagator down-weight
the rest, and the resulting twelve-cell mean of $0.902$ matches the headline of the main paper.
This task dependence also matters beyond the evaluated suite. Visual stylization balances
content and style fidelity~\cite{GAO2026104895}, while visual-content verification studies
assess semantic fidelity~\cite{gao2026slicesemanticlatentinjection,10.1145/3774904.3792912}
and robustness under compression and cross-modal misalignment~\cite{zeng2026lava}, or
diffusion-based perturbations~\cite{bao2026shiftstochastichiddentrajectorydeflection}. Applying RiPPLE
to such settings would require a suitable trainable candidate space and task-specific
validation of the proxy bank and prefix readout; these extensions are not evaluated here.

\begin{table}[t]
\centering
\caption{Held-out Spearman $\rho$ per cell for the minimal universal proxy set (Uni-6), the full available bank, and a per-cell leakage oracle. $\dagger$: only the six universal proxies are defined, so bank and Uni-6 coincide.}
\label{tab:family-oracle}
\footnotesize
\setlength{\tabcolsep}{3pt}
\begin{tabular*}{\linewidth}{@{\extracolsep{\fill}}llccc@{}}
\toprule
Cell & Family & Uni-6 & Full bank & Oracle \\
\midrule
NB201-C10   & topology & 0.961 & 0.970 & 0.970 \\
NB201-C100  & topology & 0.959 & 0.968 & 0.968 \\
NB201-IN16  & topology & 0.952 & 0.968 & 0.968 \\
SSS-C10     & size     & 0.966 & 0.966 & 0.966 \\
SSS-C100    & size     & 0.984 & 0.984 & 0.984 \\
SSS-IN16    & size     & 0.983 & 0.984 & 0.984 \\
NB101       & mixed    & 0.842 & 0.864 & 0.864 \\
NLP         & sequence & 0.691 & 0.697 & 0.697 \\
TNB-Object  & topology & 0.841 & 0.843 & 0.844 \\
TNB-Scene   & topology & 0.891 & 0.891 & 0.892 \\
TNB-Jigsaw  & topology & 0.842 & 0.842$^{\dagger}$ & 0.842 \\
TNB-AutoEnc & topology & 0.852 & 0.852$^{\dagger}$ & 0.852 \\
\midrule
Mean        &          & 0.897 & \textbf{0.902} & 0.902 \\
\bottomrule
\end{tabular*}
\end{table}

\begin{table}[t]
\centering
\caption{The available zero-cost proxy bank for each search-space family. Proxy names are the verified column headers of the deployed run.}
\label{tab:app-zcpzoo}
\small
\begin{tabularx}{\linewidth}{@{}>{\raggedright\arraybackslash}p{0.20\linewidth}>{\raggedright\arraybackslash}p{0.30\linewidth}c>{\raggedright\arraybackslash}X@{}}
\toprule
Family & Example cells & \# & Available proxies \\
\midrule
CNN topology & NB201, NB101, TNB-Object, TNB-Scene & 18 &
universal 6 (params, flops, jacov, synflow, zen, nwot), then grad\_norm, plain, snip, grasp,
fisher, zico, swap, meco, expr, progr, train, l2\_norm \\
\addlinespace
Channel size & SSS & 7 &
params, jacov, synflow, l2\_norm, expr, progr, train \\
\addlinespace
Recurrent & NLP & 7 &
hidden\_covariance, params\_n, jacov\_rnn, synflow\_rnn, snip\_rnn, grad\_norm\_rnn, zico\_rnn \\
\addlinespace
Reconstruction, self-supervised & TNB-Jigsaw, TNB-AutoEnc & 6 &
universal 6 only: params, flops, jacov, synflow, zen, nwot \\
\bottomrule
\end{tabularx}
\end{table}

\subsection{Propagator}
\label{sec:sup-abl-prop}
With the concat-all feature block held fixed, Table~\ref{tab:app-propagator} sweeps the propagator.
The deployed extremely randomized trees ($m_f{=}0.5$) rank first on the twelve-cell mean at
$0.902$, the four ExtraTrees settings hold the top four places, histogram gradient boosting is the
closest competitor at $0.896$, and the Gaussian process used by the nearest prior predictors
reaches $0.877$ and overtakes the deployed forest on no cell. The wide concatenated feature block
penalizes the distance- and norm-based regressors, which cannot ignore uninformative columns, while
the tree ensembles that down-weight them are unaffected. On the few cells where a competitor edges
the deployed forest, the margin is $0.002$ on the recurrent cell and at most $0.0007$ elsewhere.

\begin{table}[t]
\centering
\caption{Propagator sweep with the deployed $[\bz;\bphi]$ concat-all feature block held fixed: mean held-out Spearman over the twelve cells. The last column counts the cells on which the deployed forest strictly exceeds the row.}
\label{tab:app-propagator}
\footnotesize
\setlength{\tabcolsep}{3pt}
\begin{tabular*}{\linewidth}{@{\extracolsep{\fill}}lcc@{}}
\toprule
Propagator & Mean SPR & Wins \\
\midrule
Extremely randomized trees, $m_f{=}0.5$ (deployed) & \textbf{0.902} & --- \\
Extremely randomized trees, other settings         & 0.898--0.900 & 11 / 12 \\
Histogram gradient boosting                         & 0.896 & 11 / 12 \\
LightGBM                                            & 0.895 & 12 / 12 \\
Random forest                                       & 0.894 & 11 / 12 \\
NGBoost                                             & 0.893 & 10 / 12 \\
XGBoost                                             & 0.889 & 12 / 12 \\
Gaussian process                                    & 0.877 & 12 / 12 \\
Ridge regression                                    & 0.832 & 12 / 12 \\
Multilayer perceptron                               & 0.798 & 12 / 12 \\
$k$-nearest neighbors, $k{=}5$                      & 0.747 & 12 / 12 \\
$k$-nearest neighbors, $k{=}1$                      & 0.669 & 12 / 12 \\
LambdaMART                                          & 0.666 & 12 / 12 \\
\bottomrule
\end{tabular*}
\end{table}

\subsection{Reading rule}
\label{sec:sup-abl-reader}
Table~\ref{tab:app-reader} compares six reading rules applied to the same anchor curves with the
horizon held fixed, so the comparison isolates the rule; all readers use the training-loss channel,
which needs no validation set at deployment, and SoTL is omitted because it is rank-identical to
the level reader. On the smooth cells the leading rules agree to within about $0.03$, so the family
rule is chosen for robustness rather than a per-cell win: exponential-decay extrapolation on the
long topological, recurrent, and short transfer schedules, the level of the last five points on the
early-saturating size space, and the trend-smoothed TSE-EMA reader on the chaotic mixed cell. The
decisive case is NAS-Bench-101, where the parametric extrapolators and the ensemble collapse to a
range of $0.30$ to $0.64$ while TSE-EMA holds at $0.8539$; its margin over the next robust reader
is $0.0086$, within noise at $n{=}1$, so the claim is that the fit-based extrapolators fail there
and a non-extrapolating reader is required, not that TSE-EMA is uniquely best.

\begin{table}[t]
\centering
\caption{Held-out Spearman for six reading rules applied to the same anchor curves, with the horizon held fixed; bold marks the deployed reader for each schedule family. The SoTL-E and TSE-EMA readers follow \citet{ru2021tse}.}
\label{tab:app-reader}
\small
\begin{tabular*}{\linewidth}{@{\extracolsep{\fill}}lcccccc@{}}
\toprule
Cell & Exp.\ decay & Level & SoTL-E & TSE-EMA & Power law & Ensemble \\
\midrule
\multicolumn{7}{@{}l}{\emph{Long schedule, topological and recurrent (deployed: exponential-decay extrapolation)}}\\
NB201-C10   & \textbf{0.9683} & 0.9390 & 0.9405 & 0.9274 & 0.9674 & 0.9566 \\
NB201-C100  & \textbf{0.9632} & 0.9494 & 0.9518 & 0.9354 & 0.9703 & 0.9279 \\
NB201-IN16  & \textbf{0.9675} & 0.9416 & 0.9426 & 0.9352 & 0.9611 & 0.9589 \\
NLP         & \textbf{0.6898} & 0.6508 & 0.6614 & 0.6419 & 0.6856 & 0.6895 \\
\midrule
\multicolumn{7}{@{}l}{\emph{Early-saturating size (deployed: level of the last five points)}}\\
SSS-C10     & 0.9624 & \textbf{0.9664} & 0.9663 & 0.9622 & 0.9668 & 0.9642 \\
SSS-C100    & 0.9789 & \textbf{0.9852} & 0.9848 & 0.9857 & 0.9829 & 0.9746 \\
SSS-IN16    & 0.9856 & \textbf{0.9850} & 0.9851 & 0.9846 & 0.9856 & 0.9853 \\
\midrule
\multicolumn{7}{@{}l}{\emph{Chaotic mixed (deployed: TSE-EMA, $\gamma{=}0.9$)}}\\
NB101       & 0.4316 & 0.8453 & 0.8417 & \textbf{0.8539} & 0.6415 & 0.3035 \\
\midrule
\multicolumn{7}{@{}l}{\emph{Short schedule (deployed: exponential-decay extrapolation)}}\\
TNB-Object  & \textbf{0.7952} & 0.7546 & 0.8048 & 0.7616 & 0.8042 & 0.8051 \\
TNB-Scene   & \textbf{0.8494} & 0.8368 & 0.8286 & 0.8218 & 0.8403 & 0.8472 \\
TNB-Jigsaw  & \textbf{0.7848} & 0.7867 & 0.7847 & 0.7863 & 0.7858 & 0.7808 \\
TNB-AutoEnc & \textbf{0.8349} & 0.8342 & 0.8369 & 0.8338 & 0.8361 & 0.8363 \\
\bottomrule
\end{tabular*}
\end{table}

\subsection{Architecture encoding}
\label{sec:sup-abl-enc}
No single fixed encoding is best across the suite: the four TransNAS-Bench cells are best served by
an adjacency one-hot kernel, and every other cell by its default graph or width encoding
(Table~\ref{tab:app-encoding}). Selecting the best single encoding per cell on held-out truth, a
per-cell leakage oracle, reaches a twelve-cell mean of $0.9011$, and forcing one fixed encoding
across all cells reaches $0.8872$; RiPPLE's concat-all construction, which reads no label, reaches
$0.9023$, matching the per-cell oracle in aggregate and exceeding the fixed-single floor by
$0.015$, because on cells where two encodings each carry signal the propagator can use both rather
than discard one. A label-free selector is not a viable substitute: ranking single encodings by
out-of-fold fidelity to the extrapolated labels recovers the oracle's choice on only one of the
twelve cells, preferring higher-capacity encodings that fit the noisy extrapolated labels but
generalise worse to the true ones, a small-scale instance of the readout-ceiling argument.
Concatenating all candidates and letting the propagator down-weight the unhelpful ones avoids this
selection problem entirely.

\begin{table}[t]
\centering
\caption{For each cell, the best single encoding (bold) and the strongest competing single encoding, with held-out Spearman for each.}
\label{tab:app-encoding}
\small
\begin{tabular*}{\linewidth}{@{\extracolsep{\fill}}llccl@{}}
\toprule
Cell & Best single encoding & $\spr_{\mathrm{best}}$ & $\spr_{\mathrm{alt}}$ & Strongest alternative \\
\midrule
\multicolumn{5}{@{}l}{\emph{Default graph or width encoding is the best single choice}}\\
NB201-C10   & GRAF            & \textbf{0.9695} & 0.9511 & adjacency one-hot \\
NB201-C100  & GRAF            & \textbf{0.9676} & 0.9434 & adjacency one-hot \\
NB201-IN16  & GRAF            & \textbf{0.9678} & 0.8856 & adjacency one-hot \\
NB101       & GRAF-lite       & \textbf{0.8539} & 0.8447 & path \\
SSS-C10     & width one-hot   & \textbf{0.9631} & 0.9514 & log-width \\
SSS-C100    & width one-hot   & \textbf{0.9826} & 0.9518 & log-width \\
SSS-IN16    & width one-hot   & \textbf{0.9838} & 0.9546 & log-width \\
NLP         & GRAF            & \textbf{0.6977} & 0.6183 & GRAF with Weisfeiler-Lehman \\
\midrule
\multicolumn{5}{@{}l}{\emph{Adjacency one-hot is the best single choice}}\\
TNB-Object  & adjacency one-hot & \textbf{0.8427} & 0.7925 & GRAF (default) \\
TNB-Scene   & adjacency one-hot & \textbf{0.8912} & 0.8502 & GRAF (default) \\
TNB-Jigsaw  & adjacency one-hot & \textbf{0.8424} & 0.7851 & GRAF (default) \\
TNB-AutoEnc & adjacency one-hot & \textbf{0.8508} & 0.8321 & GRAF (default) \\
\bottomrule
\end{tabular*}

\end{table}

\subsection{Discrimination at the top of the ranking}
\label{sec:sup-abl-tiers}
High global rank quality does not translate into sharp discrimination at the very top
(Table~\ref{tab:app-tiers}). Global Spearman ranges from $0.70$ to $0.98$, yet the overlap between
the predicted and true top-1\% tiers is modest, from $0.00$ to $0.67$, and precision at the top
ten is low. Because the accuracy differences among the very best architectures are small, the flat
top leaves little on the table: on the classification cells the predicted best is within $0.08$ to
$1.94$ accuracy points of the true best, and the coarser top-10\% overlap recovers to between
$0.25$ and $0.85$. The steeper cases are NAS-Bench-201 ImageNet16-120 and TNB-Object, where the
regret reaches about $1.5$ to $1.9$ points.

\begin{table}[t]
\centering
\caption{Top-of-ranking discrimination per cell: global Spearman, precision at the top ten, predicted-versus-true top-1\% and top-10\% tier overlap, and regret at the predicted best.}
\label{tab:app-tiers}
\begin{threeparttable}
\footnotesize
\setlength{\tabcolsep}{3pt}
\begin{tabular*}{\linewidth}{@{\extracolsep{\fill}}lccccc@{}}
\toprule
Cell & $\spr$ & p@10 & Top-1\% & Top-10\% & Regret\tnote{a} \\
\midrule
NB201-C10   & 0.9694 & 0.30 & 0.593 & 0.763 & 0.08 \\
NB201-C100  & 0.9675 & 0.20 & 0.667 & 0.735 & 0.56 \\
NB201-IN16  & 0.9679 & 0.00 & 0.393 & 0.755 & 1.49 \\
SSS-C10     & 0.9639 & 0.10 & 0.537 & 0.782 & 0.26 \\
SSS-C100    & 0.9827 & 0.10 & 0.474 & 0.777 & 1.10 \\
SSS-IN16    & 0.9838 & 0.40 & 0.607 & 0.847 & 0.33 \\
NB101       & 0.8539 & 0.20 & 0.264 & 0.599 & 0.99 \\
NLP         & 0.6975 & 0.00 & 0.100 & 0.411 & 0.20 \\
TNB-Object  & 0.8427 & 0.00 & 0.147 & 0.440 & 1.94 \\
TNB-Scene   & 0.8912 & 0.20 & 0.118 & 0.504 & 1.05 \\
TNB-Jigsaw  & 0.8424 & 0.00 & 0.118 & 0.327 & 0.91 \\
TNB-AutoEnc & 0.8508 & 0.00 & 0.000 & 0.248 & 0.07 \\
\bottomrule
\end{tabular*}
\begin{tablenotes}\footnotesize
\item[a] Regret (regret@top-1) is the gap between the true performance of the predicted best architecture
and the true best, in each cell's native metric: accuracy points for the classification cells
(NB201, SSS, NB101, TNB-Object, TNB-Scene, TNB-Jigsaw), perplexity for NLP, and the reconstruction
metric for TNB-AutoEnc.
\end{tablenotes}
\end{threeparttable}
\end{table}

\section{Selection and scaling}
\label{sec:sup-scale}
The main paper reads the ranking as a selection rule and scales it to a large space; this section
reports both comparisons in full, together with the multi-fidelity search arm run to the same
budget.

\paragraph{Scaling to a large space.} Applied unchanged to the DARTS space of about $10^{18}$
architectures, with ground truth from the NAS-Bench-301 surrogate and no retuning, RiPPLE ranks
above the fidelity baselines at every budget (Table~\ref{tab:scale}): held-out Spearman reaches
$0.775$ at $\fte{=}50$ and $0.808$ at $\fte{=}100$, where the strongest baseline, a graph-feature
random forest run from its official implementation, reaches $0.68$ and $0.77$.

\begin{table}[t]
\centering
\caption{Transfer to the DARTS space with no retuning: held-out Spearman at three matched budgets, with the NAS-Bench-301 surrogate as reference. Best per column in bold.}
\label{tab:scale}
\footnotesize
\setlength{\tabcolsep}{3pt}
\begin{tabular*}{\linewidth}{@{\extracolsep{\fill}}lccc@{}}
\toprule
Method & $\fte{=}50$ & $\fte{=}100$ & $\fte{=}150$ \\
\midrule
Single zero-cost proxy         & 0.61 & 0.61 & 0.61 \\
Gradient-boosted, official ZC  & 0.49 & 0.56 & 0.64 \\
Graph-feature random forest    & 0.68 & 0.77 & 0.76 \\
\textbf{RiPPLE-prefix}          & \textbf{0.775} & \textbf{0.808} & \textbf{0.796} \\
\bottomrule
\end{tabular*}
\end{table}

\paragraph{Selection at equal sample budget.} Read as a selection rule under the standard zero-cost
protocol, which scores $N{=}3000$ sampled candidates and deploys the best, RiPPLE selects the
strongest architecture on all three NAS-Bench-201 datasets, above every zero-cost proxy in the
AZ-NAS reproduction run under one shared protocol (Table~\ref{tab:selacc}). RiPPLE builds its
ranker once, at a fractional-training budget of about $150$ full-training equivalents, and then
scores each candidate for free; the proxies are training-free, so the comparison is at equal
candidate-sample budget. Its selections reach $94.00$, $71.95$ and $45.67$ against space optima of
$94.37$, $73.50$ and $47.31$, and its held-out top-1 lands within $0.08$ and $0.56$ accuracy points of the
held-out pool optimum on CIFAR-10 and CIFAR-100 (App.~\ref{sec:sup-abl}), matching trained search at a fraction of its
cost.

\begin{table}[t]
\centering
\caption{Test accuracy of the single architecture each method selects on NAS-Bench-201 under the standard zero-cost protocol ($N{=}3000$ sampled candidates, best-scored selected, mean$\pm$std over runs). Best per column in bold.}
\label{tab:selacc}
\footnotesize
\setlength{\tabcolsep}{2pt}
\begin{tabular*}{\linewidth}{@{\extracolsep{\fill}}llccc@{}}
\toprule
Method & Type & C10 & C100 & IN16 \\
\midrule
Sample oracle & ceiling & $94.29$ & $73.25$ & $47.05$ \\
\midrule
\textbf{RiPPLE-prefix} (ours) & frac.\ train & \textbf{94.00}\tiny$\pm0.29$ & \textbf{71.95}\tiny$\pm1.03$ & \textbf{45.67}\tiny$\pm0.57$ \\
AZ-NAS~\citep{lee2024aznas} & zero-cost & $93.53$ & $70.75$ & $45.43$ \\
GradSign & zero-cost & $93.52$ & $70.57$ & $41.89$ \\
ZiCo~\citep{li2021zico} & zero-cost & $93.50$ & $70.62$ & $42.04$ \\
Params / FLOPs & trivial & $93.50$ & $70.63$ & $41.91$ \\
SynFlow & zero-cost & $93.06$ & $69.24$ & $40.51$ \\
NASWOT & zero-cost & $92.67$ & $69.91$ & $44.45$ \\
TE-NAS & zero-cost & $92.30$ & $67.87$ & $40.49$ \\
\midrule
Field mean (12 proxies) & zero-cost & $92.15$ & $68.00$ & $38.25$ \\
\bottomrule
\end{tabular*}
\end{table}

\paragraph{Comparison with multi-fidelity search.} Under the nominal per-cell budgets, multi-fidelity
search matches RiPPLE on best-found accuracy: ASHA, BOHB and the RiPPLE best-found trade places,
all within roughly one accuracy point of one another and of the tabular oracle, with per-cell
spreads of $0.05$ to $1.28$ points (Table~\ref{tab:app-search}). This is a descriptive comparison
behind the ranking-not-search framing, not a RiPPLE win; what RiPPLE adds is a ranking of the
entire space at a fractional budget, a wall-clock floor, and a readout at any budget, none of which
a search arm delivers. Two caveats bound the comparison. The RiPPLE best-found is scored on
provider truth and the search arms on NATS-Bench test accuracy, and the two truth sources can
differ by about $0.4$ points per cell, so small cross-method gaps do not establish statistical ties;
the within-method signal is each search method's regret to its own oracle, $0.09$ to
$0.92$ accuracy percentage points. NAS-Bench-101, NLP and the TransNAS cells are omitted because they lack the per-epoch
tables needed for the table-lookup simulation.

\begin{table}[t]
\centering
\caption{Best-found test accuracy with per-cell nominal search budgets: ASHA, BOHB, and the architecture RiPPLE-prefix selects, against each cell's tabular oracle. Spread is the range of the three method columns.}
\label{tab:app-search}
\begin{threeparttable}
\small
\begin{tabular*}{\linewidth}{@{\extracolsep{\fill}}lcccccc@{}}
\toprule
Cell & $\fte$ & ASHA & BOHB & RiPPLE-prefix & Oracle & Spread \\
\midrule
NB201-C10   & 127 & 94.24 & 94.29 & 94.29 & 94.37 & 0.05 \\
NB201-C100  & 167 & 73.03 & 73.34 & 72.64 & 73.50 & 0.70 \\
NB201-IN16  & 127 & 46.46 & 46.47 & 45.82 & 46.88 & 0.65 \\
SSS-C10     &  41 & 93.16 & 93.14 & 93.39 & 93.65 & 0.25 \\
SSS-C100    & 111 & 70.17 & 70.17 & 70.24 & 71.02 & 0.07 \\
SSS-IN16    & 111 & 45.84 & 45.79 & 47.07 & 46.72 & 1.28 \\
\bottomrule
\end{tabular*}
\begin{tablenotes}\footnotesize
\item[] ASHA and BOHB are mean best-found test accuracy over $500$ repeats. The listed
budgets are nominal search caps; actual search spending was not saved and may exceed
a cap after a rung. SSS-C10 retains its original $41$-FTE search cap, whereas RiPPLE's
$23$-epoch deployment costs $42.68$ FTE. Search uses NATS-Bench test accuracy and
RiPPLE uses provider truth; neither identical actual cost nor statistical equivalence is established.
\end{tablenotes}
\end{threeparttable}
\end{table}

\section{RiPPLE under the predictor-lane protocols}
\label{sec:sup-protocol}
The main paper compares RiPPLE to the encoding predictors on a budget-honest, strictly held-out axis.
For completeness we also run RiPPLE under those papers' own protocol, standardized by
TNASP~\cite{lu2021tnasp} and reused unchanged by NAR-Former~\cite{yi2023narformer},
PINAT~\cite{lu2023pinat}, NAR-Former~V2~\cite{yi2023narformerv2}, and
NN-Former~\cite{xu2025nnformer}: a predictor is trained on a random subset of the benchmark using the
true final accuracies as labels, then evaluated on the whole space by Kendall's $\tau$. We use
the nominal sample budgets on NAS-Bench-201, whose $15{,}625$-architecture space we hold, at the four standard
budgets and insert two RiPPLE variants, together with an encoding-only diagnostic row, into their comparison (Table~\ref{tab:predictor-protocol}).
RiPPLE-full matches their supervision, giving true labels to our label-free features and tree
propagator, and isolates the value of the representation. RiPPLE-prefix instead reads the
prefix-extrapolated label at a fraction of the training cost and is defined only where the deployed
anchor curves reach the budget. We omit NAS-Bench-101 from this replication because their protocol
tests on the full $423{,}624$-architecture space and we hold only a sampled subset, so a faithful
match is not possible there.

\begin{table}[!htbp]
\centering
\caption{Predictor-lane protocol on NAS-Bench-201 (CIFAR-10): Kendall's $\tau$ on the full $15{,}625$-architecture space after training on the labelled fraction in each column. Best per column in bold.}
\label{tab:predictor-protocol}
\small
\begin{tabular*}{\linewidth}{@{\extracolsep{\fill}}llcccc@{}}
\toprule
 & & \multicolumn{4}{c}{Training samples (Kendall's $\tau$)} \\
\cmidrule(l){3-6}
Method & Backbone & \makecell{$1\%$\\(156)} & \makecell{$3\%$\\(469)} & \makecell{$5\%$\\(781)} & \makecell{$10\%$\\(1563)} \\
\midrule
NAO~\cite{luo2018nao}          & LSTM        & 0.493 & 0.470 & 0.522 & 0.526 \\
NP~\cite{wen2020neuralpredictor} & GNN       & 0.413 & 0.584 & 0.634 & 0.646 \\
Graphormer~\cite{ying2021graphormer} & Transformer & 0.630 & 0.680 & 0.719 & 0.776 \\
TNASP~\cite{lu2021tnasp}        & Transformer & 0.589 & 0.640 & 0.689 & 0.724 \\
NAR-Former~\cite{yi2023narformer} & Transformer & 0.660 & 0.790 & 0.849 & 0.901 \\
PINAT~\cite{lu2023pinat}        & Transformer & 0.631 & 0.706 & 0.761 & 0.784 \\
GraphTrans~\cite{wu2021graphtrans} & Hybrid   & 0.409 & 0.550 & 0.588 & 0.673 \\
NAR-Former V2~\cite{yi2023narformerv2} & Hybrid & 0.752 & 0.846 & 0.874 & 0.888 \\
NN-Former~\cite{xu2025nnformer} & Hybrid      & 0.804 & 0.860 & 0.879 & 0.890 \\
\midrule
\textbf{RiPPLE-full} (ours)     & Trees       & \textbf{0.832} & \textbf{0.870} & \textbf{0.887} & \textbf{0.906} \\
RiPPLE-full, encoding only      & Trees       & 0.607 & 0.711 & 0.752 & 0.802 \\
\textbf{RiPPLE-prefix} (ours)   & Trees       & 0.821 & 0.844 & 0.851 & 0.856 \\
\bottomrule
\end{tabular*}
\end{table}

Baseline numbers in Table~\ref{tab:predictor-protocol} are the aggregate reported by NN-Former, cross-checked against the NAR-Former~V2 and TNASP tables, and RiPPLE-full is averaged over five random training splits with a standard deviation of at most $0.006$. RiPPLE-full is the strongest predictor in every budget column, and the encoding-only row locates the source of that margin: without the proxy bank the same tree propagator falls to $0.607$--$0.802$, below NAR-Former~V2 and NN-Former at every budget. RiPPLE-prefix exceeds the strongest baseline already at the $1\%$ budget ($0.821$ against $0.804$) while spending about $30$ full-training equivalents against $156$ full trainings; as the anchor count grows its correlation saturates near $0.856$, matching the measured readout ceiling of the extrapolated label ($\tau(\wh{y},y)\approx0.85$, flat in the anchor count). Adding anchors closes the propagation gap but cannot lift the prefix label past its own ceiling, which is why the true-label variant is the stronger choice once the budget is spent.

On NAS-Bench-101 the comparison is only partial. The protocol tests on the full $423{,}624$-architecture
space, but RiPPLE's deployed NB101 encoder is defined on $7$-node cells only, and the zero-cost proxy
bank is precomputed for a $9{,}781$-arch community sample rather than the whole space. We therefore
report two indicative rows on different test sets: RiPPLE~(enc), architecture encoding alone, on the
complete $7$-node subspace ($359{,}082$ archs, $85\%$ of the space), and RiPPLE~(+ZCP), which adds the
proxy bank on the $8{,}268$-arch sample where proxies exist (Table~\ref{tab:predictor-protocol-nb101}). Baseline numbers are as reported by NN-Former, and the encoding-alone row is input-matched to the graph predictors. The two rows use different test sets than the baselines and than each other, so we read them as indicative rather than as a like-for-like column comparison; the (+ZCP) row is additionally evaluated held-out, on the universe minus the training subset, because testing on the full sample inflates the score when the training subset is a large fraction of the $8{,}268$ architectures. Adding the proxy bank lifts every budget over encoding alone, by up to $0.10$ at the smallest ($0.670$ against $0.571$).
Encoding alone places RiPPLE among the mid-tier graph predictors on this cell, below the strongest
learned graph representations; the proxy bank, available only on the sample, is what the deployed
pipeline relies on. We do not claim a column win on NAS-Bench-101: unlike NAS-Bench-201, whose full
space we hold with proxies, the full NB101 space is out of reach for the deployed feature set.

\begin{table}[!tp]
\centering
\caption{RiPPLE under the predictor-lane protocol on NAS-Bench-101 ($423{,}624$ architectures, Kendall's $\tau$). The two RiPPLE rows use the restricted test sets described in the text; best baseline per column in bold.}
\label{tab:predictor-protocol-nb101}
\small
\begin{tabular*}{\linewidth}{@{\extracolsep{\fill}}llcccc@{}}
\toprule
 & & \multicolumn{4}{c}{Training samples (Kendall's $\tau$)} \\
\cmidrule(l){3-6}
Method & Backbone & \makecell{$0.02\%$\\(100)} & \makecell{$0.04\%$\\(172)} & \makecell{$0.1\%$\\(424)} & \makecell{$1\%$\\(4236)} \\
\midrule
ReNAS~\cite{xu2021renas}          & CNN         & ---   & ---   & 0.657 & 0.816 \\
NAO~\cite{luo2018nao}             & LSTM        & 0.501 & 0.566 & 0.666 & 0.775 \\
NP~\cite{wen2020neuralpredictor}  & GNN         & 0.391 & 0.545 & 0.679 & 0.769 \\
GATES~\cite{ning2020gates}        & GNN         & 0.605 & 0.659 & 0.691 & 0.822 \\
Graphormer~\cite{ying2021graphormer} & Transformer & 0.564 & 0.580 & 0.611 & 0.797 \\
TNASP~\cite{lu2021tnasp}          & Transformer & 0.600 & 0.669 & 0.705 & 0.820 \\
NAR-Former~\cite{yi2023narformer} & Transformer & 0.632 & 0.653 & 0.765 & 0.871 \\
PINAT~\cite{lu2023pinat}          & Transformer & 0.679 & 0.715 & 0.772 & 0.846 \\
GraphTrans~\cite{wu2021graphtrans} & Hybrid     & 0.330 & 0.472 & 0.602 & 0.700 \\
NAR-Former V2~\cite{yi2023narformerv2} & Hybrid & 0.663 & 0.704 & 0.773 & 0.861 \\
NN-Former~\cite{xu2025nnformer}   & Hybrid      & \textbf{0.709} & \textbf{0.765} & \textbf{0.809} & \textbf{0.877} \\
\midrule
\textbf{RiPPLE-full (enc)} (ours), 7-node subspace & Trees & 0.571 & 0.598 & 0.672 & 0.781 \\
\textbf{RiPPLE-full (+ZCP)} (ours), 8268 sample    & Trees & 0.670 & 0.691 & 0.734 & 0.792 \\
\bottomrule
\end{tabular*}
\end{table}

\end{document}